\documentclass[lettersize,journal, twocolumn]{IEEEtran}
\usepackage{amsthm,amsmath,amssymb}
\usepackage{mathrsfs}
\usepackage{algorithm}
\usepackage{algpseudocode}
\usepackage{array}
\usepackage[caption=false,font=footnotesize,labelfont=rm,textfont=rm]{subfig}
\usepackage{textcomp}
\usepackage{stfloats}
\usepackage{makecell}
\usepackage{url}
\usepackage{bm}
\usepackage{hyperref}
\usepackage{graphicx}
\usepackage{cite}
\usepackage{color}
\usepackage{multirow}
\hypersetup{hidelinks,
colorlinks=true,
allcolors=black,
pdfstartview=Fit,
breaklinks=true}
\usepackage[table,xcdraw,dvipsnames]{xcolor}  
\usepackage{xpatch}
\makeatletter
\def\changeBibColor#1{%
\in@{#1}{485127}
\ifin@\color{black}\else\normalcolor\fi
}

\begin{document}

\title{Exploring Text-Guided Single Image Editing for Remote Sensing Images}

\author{Fangzhou Han, Lingyu Si, Hongwei Dong,~\IEEEmembership{Member,~IEEE}, Zhizhuo Jiang, Lamei Zhang,~\IEEEmembership{Senior Member,~IEEE}, Hao Chen,~\IEEEmembership{Member,~IEEE}, Yu Liu, ~\IEEEmembership{Member,~IEEE} and Bo Du,~\IEEEmembership{Senior Member,~IEEE}
\thanks{This work was supported by National Key R\&D Program of China under Grant 2021YFA0715202, National Natural Science Foundation of China under Grants  62301539, 62271172, 62425117, the Open Fund of (No. WDZC20255290403) \textit{(Corresponding author: Hongwei Dong, Lamei Zhang)}.  }
\thanks{F. Han, L. Zhang and H. Chen are with the Department of Information Engineering, Harbin Institute of Technology, Harbin, China (e-mail: \{lmzhang, hit\_hao\}@hit.edu.cn).}
\thanks{H. Dong and L. Si are with the National Key Laboratory of Space Integrated Information System, Institute of Software, Chinese Academy of Sciences and the National Key Laboratory of Information Systems Engineering, Beijing, China (e-mail: \{donghongwei,lingyu\}@iscas.ac.cn).}
\thanks{Z. Jiang and Y. Liu are with Shenzhen International Graduate School, Tsinghua University, Shenzhen 518055, China.}
\thanks{B. Du is with the Hubei Luojia Laboratory, National Engineering Research Center for Multimedia Software, School of Computer Science, Wuhan University, Wuhan, China (e-mail: dubo@whu.edu.cn).}
\thanks{F. Han and L. Si contributed equally to this paper.}
}

\markboth{Journal of \LaTeX\ Class Files,~Vol.~14, No.~8, August~2021}{}%


\maketitle
\begin{abstract}
Artificial intelligence generative content (AIGC) has significantly impacted image generation in the field of remote sensing. However, the equally important area of remote sensing image (RSI) editing has not received sufficient attention. Deep learning based editing methods generally involve two sequential stages: generation and editing. \textcolor{black}{For natural images, these stages primarily rely on generative backbones pre-trained on large-scale benchmark datasets and text guidance facilitated by vision-language models (VLMs). However, it become less viable for RSIs: First, existing generative RSI benchmark datasets do not fully capture the diversity of RSIs, and is often inadequate for universal editing tasks. Second, the single text semantic corresponds to multiple image semantics, leading to the introduction of incorrect semantics.} To solve above problems, this paper proposes a text-guided RSI editing method and can be trained using only a single image. A multi-scale training approach is adopted to preserve consistency without the need for training on extensive benchmarks, while leveraging RSI pre-trained VLMs and prompt ensembling (PE) to ensure accuracy and controllability. Experimental results on multiple RSI editing tasks show that the proposed method offers significant advantages in both CLIP scores and subjective evaluations compared to existing methods. Additionally, we explore the ability of the edited RSIs to support disaster assessment tasks in order to validate their practicality. Codes will be released at \href{https://github.com/HIT-PhilipHan/remote_sensing_image_editing}{https://github.com/HIT-PhilipHan/remote\_sensing\_image\_editing}.    
\end{abstract}


\section{Introduction}

\textcolor{black}{\IEEEPARstart{B}{ecause} of the extensive coverage and wealth of information, remote sensing images (RSIs) have facilitated various vital tasks such as disaster response, environmental monitoring, and urban planning \cite{10380634,10542538}. Nowadays, the cost of remote sensing has significantly decreased, providing sufficient data for long-term observations of stable areas such as farmland, mountains and cities. However, in extreme events such as earthquakes, forest fires, and tsunamis, timely and comprehensive observation is challenging due to their low frequency of occurrence and high randomness in location. The difficulty in obtaining RSIs for such scenarios further hinders their prevention, detection, and assessment.}

\textcolor{black}{It is crucial to emphasize the strong correlation between extreme and conventional scenarios,} where the basic semantic of objects in RSIs remain unchanged, but new semantic information is added, such as transforming a tree into a burning tree. This allows for the acquisition of a large number of RSIs for extreme scenarios by editing RSIs from conventional ones. However, the significant importance of RSI editing has largely been overlooked in current research.


While research on image editing remains insufficient, the field of remote sensing has already seen comprehensive studies on image generation, which serves as both the foundation and the initial stage of editing. In recent years, the development of deep learning has led to extensive applications of generation networks, such as variational autoencoder (VAE) \cite{NEURIPS2018_73e5080f} and generative adversarial network (GAN) \cite{10287367,9316788}. These advancements have led to significant breakthroughs in remote sensing field, including sample augmentation \cite{9250622,9858033,10025584,9531488}, image super-resolution \cite{8677274, 10403859, 10763472}, cross-modal image transformation \cite{9323085, 10707182,10103165}, mitigating label noise \cite{10382538}, cross-domain adaptation \cite{10716600} and cloud removal \cite{9481173}. VAE generates images by sampling from the latent distribution mapped from the training images, while GAN produces sufficiently realistic image generation results through adversarial games between the generator and discriminator. Although the aforementioned methods perform well in generation tasks, their aim is to fit the distribution of existing data and do not have the ability to introduce new semantics required for editing. More importantly, the development of remote sensing platforms has significantly reduced the cost of obtaining RSIs in conventional scenarios, greatly diminishing the need for unconditional generation of RSIs.

\textcolor{black}{Unlike image generation, the core of image editing lies in introducing new semantic information into the target image, which requires guidance from additional information.} This is inherently a challenging task, but the emergence of vision-language models (VLMs) offers an excellent approach for providing this additional information \cite{10445007}. For instance, the seminal work in VLMs, contrastive language-image pre-training (CLIP), achieves semantic alignment between text and images through pre-training on large-scale paired text-image datasets \cite{radford2021learning}, which is essential for text-guided image editing. Subsequently, the integration of CLIP with various generative models, such as GANs and denoising diffusion probabilistic models (DDPMs) \cite{ho2020denoising} has led to significant advancements in controllable text-guided image editing \cite{xia2021tedigan,Patashnik_2021_ICCV}. Generally, such methods can be classified into below categories\cite{huang2024}: training-based approaches \cite{10387416, Wang_2023_ICCV}, testing-time fine-tuning approaches \cite{Mokady_2023_CVPR, Zhang_CVPR}, training and fine-tuning free approaches \cite{Wallace_2023_CVPR, NEURIPS2023_3469b211}. Training-based methods are distinguished not only by their robust training of diffusion models and their effective data distribution modeling, but also by their reliable performance across diverse editing tasks. In contrast, testing-time fine-tuning approaches offer greater precision and controllability, while fine-tuning free methods are significantly faster and more cost-efficient. While these methods have led to significant advancements in image editing performance, applying these existing models to the field of remote sensing presents numerous challenges. Specifically, deep learning based image editing models can be progressively divided into two stages: generation and editing. In the generation stage, the model produces content that maintains semantic consistency with the original image, while in the editing stage, new semantic information is introduced to fulfill the editing objectives. Next, we will analyze the potential factors within each of these two stages that contribute to the poor performance of conventional editing models on RSIs.

Firstly, during the generation stage, conventional image editing models struggle to preserve content and specific details from the original RSIs, such as background elements and the identity of objects within them \cite{Kawar_2023_CVPR}. This limitation arises because the models, when applied to natural images, rely on DDPMs trained on large and diverse benchmark datasets, which equips them to generate images across a broad spectrum of complex scenarios. However, for RSIs, the above approach no longer works because remarkable domain discrepancies often arise in terms of resolution, perspective, and other sensor-related factors between the images used for training and those intended for editing \cite{8672156}. Furthermore, the semantic information pertinent to the RSIs requiring edited may be underrepresented in the training dataset, such as uncommon ships, buildings, and newly emerged land cover types \cite{Li_2024_CVPR}. As a result, backbones trained on RSI benchmark datasets lack sufficient generalization capability required to effectively handle the diverse editing tasks needed for RSIs. 

\begin{figure}[ht] \color{black}
\centering
\includegraphics[width=1.0\linewidth]{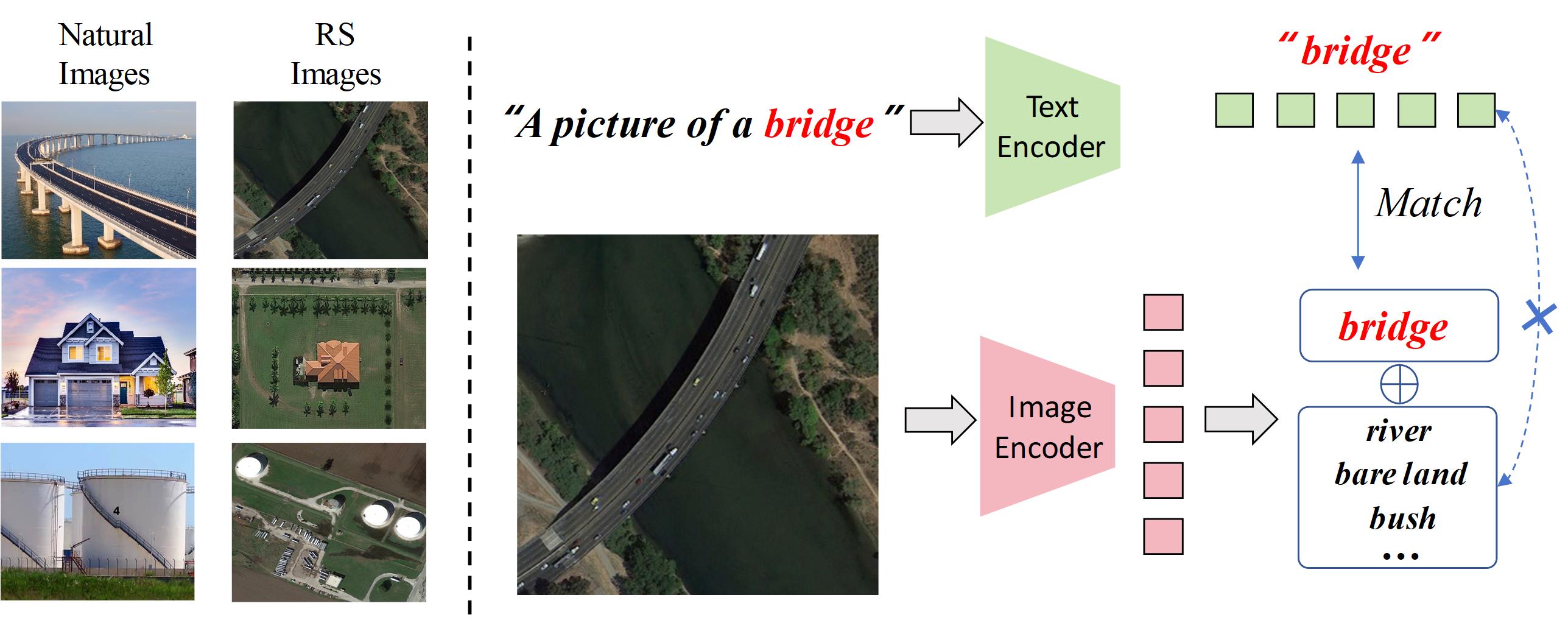}
\caption{Semantic mismatches between text and images. In the image, semantic information encompasses not just the bridge but also the river, bush, bare land, and other surrounding elements. However, the corresponding textual semantic information often refers to the most prominent part of the image, which in this case is the bridge.}
\label{fig:question2}
\end{figure}

Secondly, during the editing stage, conventional text-guided image editing models face the challenge of semantic mismatches between text and images. This issue arises from two key reasons. First, VLMs used in conventional editing methods are typically pre-trained on natural text-image pairs, limiting their capacity to effectively interpret RSIs. Second, RSIs have a much higher spatial coverage compared to natural images, often containing multiple semantic elements \cite{10356128}. However, pre-trained VLMs like CLIP typically align the entire image with a single text description, and this one-to-many correspondence leads to inaccurate semantic alignment. Fig. \ref{fig:question2} illustrates the aforementioned problem with an example. Such semantic mismatches can easily result in inaccurate results from the model during the text-guided editing process.

To tackle the aforementioned challenges, this paper proposes a text-guided image editing method customized for RSIs, with optimizations applied concurrently in both the image generation and editing stages of the RSI editing process. To address the issue of semantic inconsistency before and after editing, \textcolor{black}{the proposed method} implement a multi-scale training strategy using only the target image for DDPM training, which essentially eliminates the semantic inconsistency in the generation process caused by domain discrepancies. To address the issue of text-image semantic mismatch in the editing stage, we leverage a RSI pre-trained CLIP model to incorporate remote sensing information into image editing. Additionally, \textcolor{black}{the proposed method} introduces prompt ensembling (PE), which generates similar text through large language models (LLMs). Combining multiple texts helps align text embeddings more closely with the desired image embeddings and avoids introducing anomalous semantics. Overall, the contributions of this paper are as follows:
\begin{enumerate}
\item {We focus on the relatively overlooked but important issue of RSI editing and propose a RSI-customized method that achieves stable and controllable text-guided RSI editing.}
\item {We implement a multi-scale training strategy with a single image to solve the problem of lacking corresponding RSIs for editing tasks, thereby maintaining consistency in content and details between the original and edited images.}
\item {We leverage RSI pre-trained CLIP and propose the PE technique to address the issue of irrelevant semantic interference during editing, ensuring the accuracy of text-guided editing results.}
\item {Extensive experimental results demonstrate that the proposed method is adaptable to various RSI editing tasks.}
\end{enumerate}

\section{Related Works}

\subsection{Generation Models in Remote Sensing Field}

In recent years, with the advancement of deep generative models, RSIs generation has yielded significant results in various tasks, including data augmentation, image super-resolution, cross-modal image transformation, and cloud removal. Broadly speaking, the generative models predominantly used in the field of remote sensing are GANs and DDPMs.

\textcolor{black}{With regard to} the GANs, for the data augmentation task, Zhang \textit{et al.} \cite{9393473} proposed a data augmentation method based on variational autoencoder multi-scale GAN with spatial and channel wise attention to balance the sample distribution and improve the subsequent ROI extraction results. Chen \textit{et al.} \cite{9386248} proposed a novel data-level solution, namely, instance-level change augmentation, to generate bitemporal images that contain changes involving plenty and diverse buildings by leveraging generative adversarial training. Then, they proposed context-aware blending for a realistic composite of the building and the background. For the image super resolution task, Dong \textit{et al.} \cite{9328132} explored the potential of the reference-based super-resolution method on RSIs, utilizing rich texture information from high resolution reference images to reconstruct the details in low resolution images. 


Recently, DDPM-based models have also gained traction in the field of remote sensing image generation, particularly in super-resolution tasks. For instance, Xiao \textit{et al.} \cite{10353979} introduced a DDPM-based approach for RSI super-resolution called EDiffSR. EDiffSR was not only easy to train but also retained the advantages of DDPM in generating perceptually pleasing images. Experiments on remote sensing datasets demonstrated that EDiffSR could effectively restore visually appealing images in both simulated and real-world RSIs. 


\textcolor{black}{To sum up, although the generation of remote sensing image has been fully studied, the importance of remote sensing image editing has been neglected. Different from the unconditional image generation, the core of remote sensing image editing is to introduce new semantic information into the target image through the guidance of the additional information.The mentioned models are mainly concerned with the reproduction of the original information of the image, so it cannot meet this challenge, which is the motivation behind our proposed approach.}

\subsection{Denoising Diffusion Probabilistic Model (DDPM)}

In the process of diffusion, DDPMs convert the original data into a normal distribution, which means that every point on the normal distribution is a map of the real data, thus enhancing the interpretability of the model. Meanwhile, compared with GANs, the diffusion model mitigates the problems such as collapse during training, and the quality of images generated is improved. In recent years, research on the structure, optimization goals and application scenarios of DDPM model has become increasingly abundant. For example, as to the model structure and the optimization goals, Peebles \textit{et al.} \cite{Peebles_2023_ICCV} explored a new class of diffusion models based on the transformer architecture. The model trained latent diffusion models of images, replacing the commonly used U-Net backbone with a transformer that operates on latent patches. Kingma \textit{et al.} \cite{NEURIPS2021_b578f2a5} introduced a family of diffusion-based generative models that obtain state-of-the-art likelihoods on standard image density estimation benchmarks. It showed that the variational lower bound simplifies to a remarkably short expression in terms of the signal-to-noise ratio of the diffused data, improving the theoretical understanding of this model class. Kang \textit{et al.} \cite{10633292} applied the conditional diffusion model low-light image enhancement (LIE) to restore the details in dark regions. Liu \textit{et al.} \cite{10480695} proposed VADiffusion, a compressed domain information guided conditional diffusion framework to identify sudden anomalies and detect persistent anomalies in video. \textcolor{black}{Kulikov \textit{et al.} \cite{pmlr-v202-kulikov23a} introduced a framework for training a DDM on a single image. The method, which coined SinDDM, learns the internal statistics of the training image by using a multi-scale diffusion process. To drive the reverse diffusion process, they use a fully-convolutional light-weight denoiser, which is conditioned on both the noise level and the scale. This architecture allows generating samples of arbitrary dimensions, in a coarse-to-fine manner.} 



\textcolor{black}{In general, in order for DDPM to have the ability to generate images from noise, it is necessary to train for a long time on a large dataset. In addition to the consumption of computing resources, it also has drawbacks for image editing tasks. That is, generative models trained on large-scale data sets cannot guarantee the consistency of image content before and after editing. In our proposed method, we only need to train DDPM through a single remote sensing image, thus solving the above problems and greatly reducing the dependence on large-scale datasets.}

\subsection{Image Editing Methods}

In the field of AIGC, which utilizes artificial intelligence to create and modify digital content, image editing is considered an important area for innovation and practical applications \cite{huang2024,10542240}. Generally, such image editing methods can be classified into below categories: training-based approaches, testing-time fine-tuning approaches, training and fine-tuning free approaches. \textcolor{black}{With regard to} the training-based approaches, Kawar \textit{et al.} \cite{Kawar_2023_CVPR} demonstrated, for the very first time, the ability to apply complex (e.g., non-rigid) text-based semantic edits to a single real image named Imagic. Yang \textit{et al.} \cite{10154005} presented a novel approach called dual-cycle diffusion, which generates an unbiased mask to guide image editing. For the testing-time fine-tuning approaches, these approaches range from fine-tuning the entire denoising model to focusing on specific layers or embeddings. For example, Mokady \textit{et al.} \cite{Mokady_2023_CVPR} introduced an accurate inversion technique and thus facilitate an intuitive text-based modification of the image. They only modify the unconditional textual embedding that is used for classifier-free guidance, rather than the input text embedding. Finally, for the training and fine-tuning free approaches, they are fast and low-cost because they do not require any form of training (on the dataset) or fine-tuning (on the source image) during the complete process of editing. For example, Wallace \textit{et al.} \cite{Wallace_2023_CVPR} proposed exact diffusion inversion via coupled transformations (EDICT), an inversion method that drawn inspiration from affine coupling layers.

\textcolor{black}{Although diffusion-based image editing models have garnered widespread attention in fields such as medical image processing and computer vision, their potential remains unexplored in the domain of remote sensing. As far as we know, the proposed method is the first try of exploring the text-guided editing approach for remote sensing images.}

\section{Methodology}

\textcolor{black}{This section elaborates on our proposed framework for text guided single image editing for RSIs.  \textcolor{black}{It consists of two key components, i.e., the multi-scale training strategy based on single image and the prompt-guided fine-tuning process, which received inspiration from the sinDDM \cite{pmlr-v202-kulikov23a}.} The proposed approach firstly train the SinDDM on just one single image, then the trained SinDDM is fine-tuned using different types of prompt guidance such as text and ROIs. Different from the existing image editing methods, the proposed method solves the problem of lacking numerous corresponding RSIs for editing tasks, thereby maintaining consistency in content and details between the original and edited images. Moreover, the proposed editing strategy employ PE as a pre-processing strategy to produce robust text guidance, mitigating the mismatch between text and image embeddings.}

\subsection{Multi-scale training strategy based on single image}

The training process of conventional DDPMs typically requires a large number of samples to ensure the effective generation of images from noise. Although extensive benchmark datasets provide DDPMs with generative capabilities, remote sensing images exhibit significant variations in resolution, color, and other factors across different sensors. This results in inadequate generalization for backbones trained on such datasets, causing semantic inconsistencies between the original and edited RSIs. To resolve this, \textcolor{black}{the proposed method} introduces a multi-scale DDPM training strategy, termed SinDDM \cite{pmlr-v202-kulikov23a}, which operates with a single image, maintaining consistency in content and details between the original and edited RSIs.

\begin{figure}[!ht] \color{black}
\centering
\includegraphics[width =0.5\textwidth]{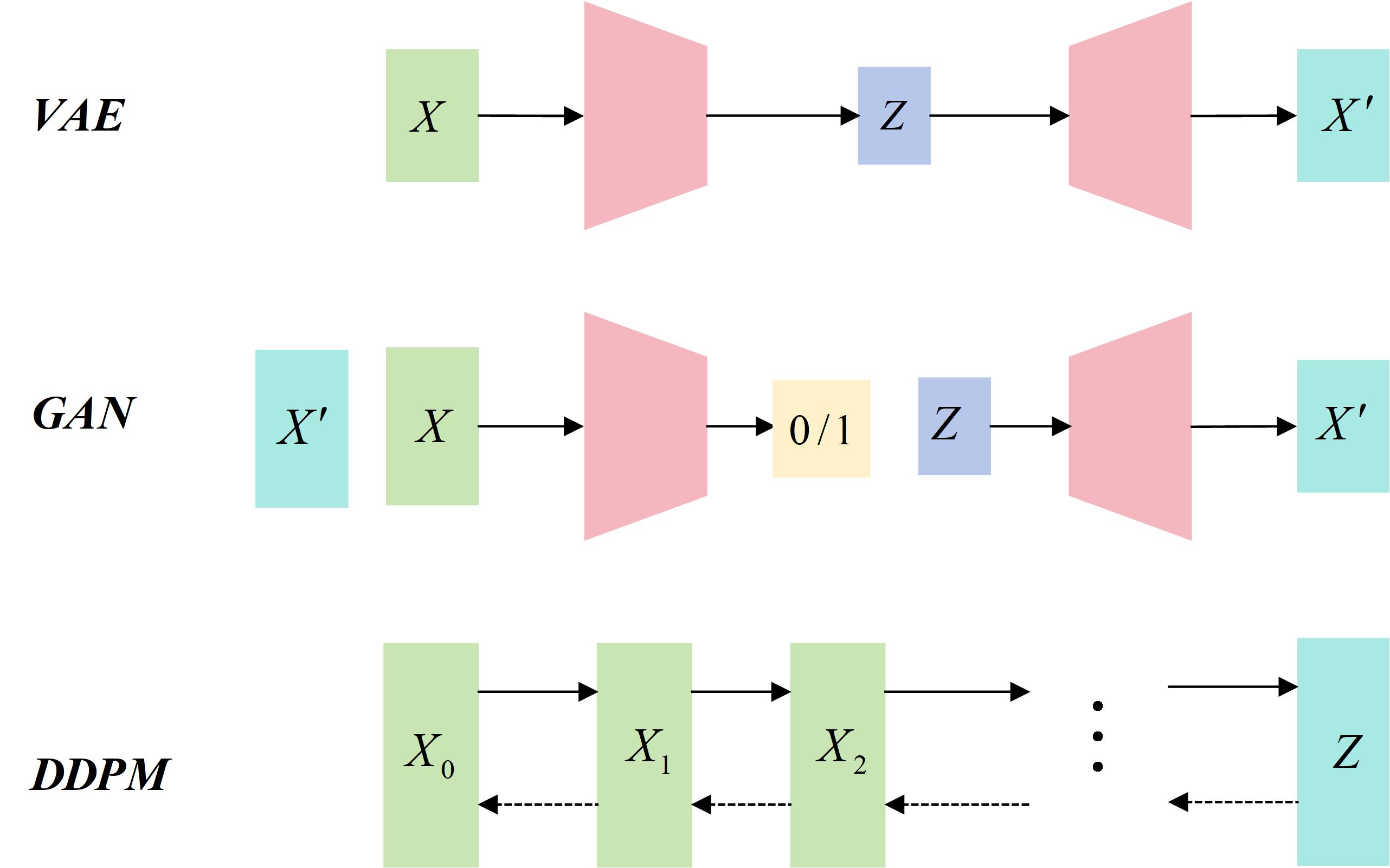}
\caption{The structure comparison of the VAE, GAN and the DDPM.}
\label{models}
\end{figure}

\begin{figure*}[!ht]
\centering
\includegraphics[width = 1.0\textwidth]{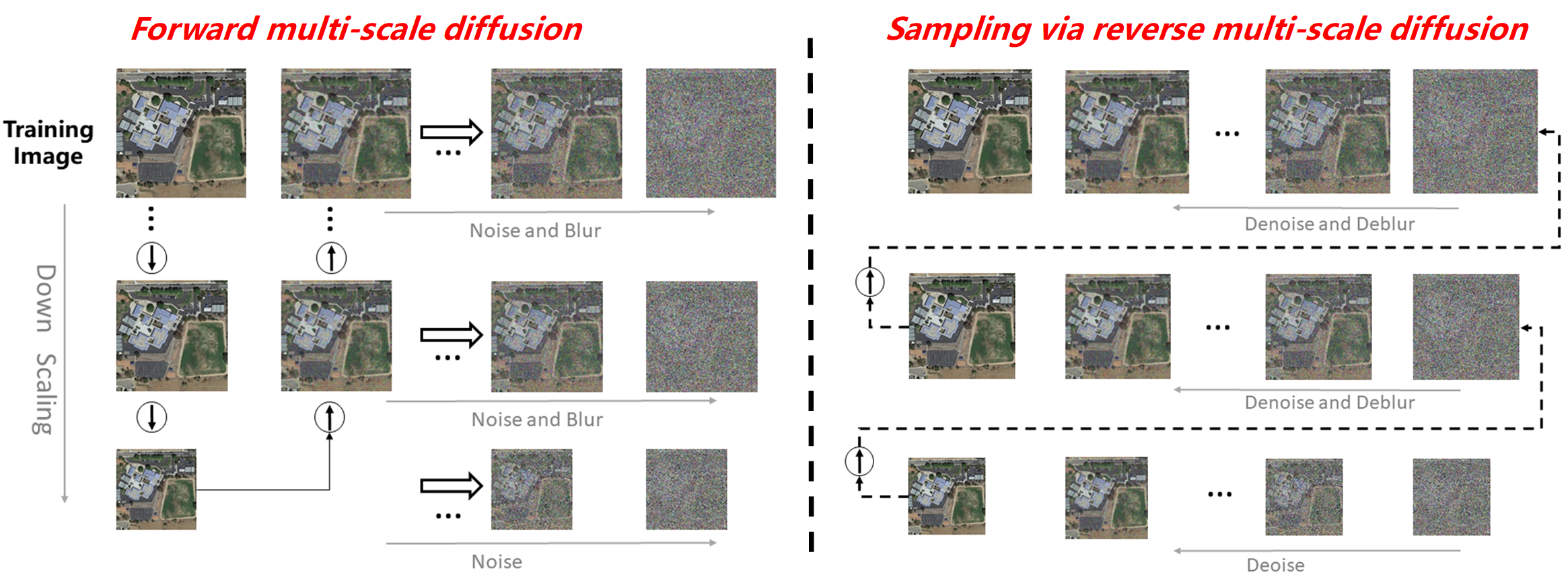}
\caption{The multi-scale training strategy. The forward multi-scale diffusion process (left) construct a sequence of images that are linear combinations of the original images in multi scale, their blurry version, and noise. For the sampling process via reverse multi-scale diffusion (right) starts from pure noise at the coarsest scale. In each scale, SinDDM gradually removes the noise until reaching a clean image, which is then upsampled and combined with noise to start the process again in the next scale. It is worth noting that the above process uses a multi-scale training strategy, so just a single image could achieve effective training of DDPM.}
\label{pretrain}
\end{figure*}

We describe SinDDM by starting with the conventional DDPM training process. The idea behind DDPM is to define a Markov chain with diffusion steps to incrementally introduce random noise to the data, and subsequently learning to reverse this diffusion process to reconstruct the desired image from the noise. Therefore, the DDPM can be divided into two process: forward process and backward process. The structure comparison of DDPM, VAEs and GANs is shown in the Fig. \ref{models}. SinDDM is consistent with the conventional DDPM in the construction of denoising model while the difference is that SinDDM constructs multi-scale pyramid by down-sampling the input single image in the training process, and trains the denoising model since the smallest scale. Upon completion of denoising at a specific scale, the process transitions to denoising at the subsequent scale through up-sampling. Therefore, in the original DDPM, there is only one parameter $T$, representing the number of noise addition steps in the forward process, while in the SinDDM, a new parameter $S$ is added to denote the scale within the denoising model. The whole multi-scale diffusion is shown in Fig. \ref{pretrain}.

First, we could construct the image pyramid $\text{ }\!\!\{\!\!\text{ }{{x}^{N-1}},{{x}^{N-2}},\cdots ,{{x}^{0}}\text{ }\!\!\}\!\!\text{ }$ while every image ${x}^{s}$ of the pyramid is generated via the down-sampling factor $r$. We use bicubic interpolation for both the upsampling and downsampling operations. Then, for one scale $s\in (0,1,\cdots ,N-1)$, the forward process $q(x_{t}^{s}|x_{t-1}^{s})$ could be expressed by the following formula:
\begin{equation}
q(x_{t}^{s}|x_{t-1}^{s})=\mathcal{N}(x_{t}^{s};\sqrt{1-\beta _{t}}x_{t-1}^{s},\beta _{t}\mathit{I})
\end{equation}
where the $t$ represents the steps of forward process to add noise to the image, which grows from $0$ to $T$ and the $\beta_{t}$ is the variance taken at each step, which is between 0 and 1. we use $\alpha_{t}$ to replace $1-\beta_{t}$ and DDPM uses a linear variance schedule, modeling the denoising process a Markov chain:
\begin{equation}
q(x_{1:T}^{s}|{{x}})=\prod\limits_{t=1}^{T}{q(x_{t}^{s}|{{x}^{s}})}.
\end{equation}

Then through reparameterization trick, the noise image $x_{t}^{s}$ of any step can be sampled directly based on the original data $x^{s}$:
\begin{equation}
\begin{split}
    &{{\bar{\alpha }}_{t}}=\prod\limits_{i=1}^{t}{{{{\bar{\alpha }}}_{t}}} \\
    x_{t}^{s}=\sqrt{{{{\bar{\alpha }}}_{t}}}{{x}^{s}}&+\sqrt{1-{{{\bar{\alpha }}}_{t}}}\epsilon,\quad while\text{ }\epsilon \sim \mathcal{N}(\boldsymbol{\mathit{0,I}})
\end{split}
\end{equation}
where, the $\epsilon$ is the noise. When step grows, the ${{\bar{\alpha }}_{t}}$ decreases monotonically from $1$ to $0$, which ensures that the resulting image $x_T$ is close to random noise.

It should be noted that due to the upsampling and downsampling operations based on interpolation, the generated image will be blurred, so it is necessary to achieve de-blurring while de-noising in the training process. To do this, we blend the noisy image with its blurry version. We construct a blurry version of the pyramid $\text{ }\!\!\{\!\!\text{ }{{x}^{N-1}},{{x}^{N-2}},\cdots ,{{x}^{0}}\text{ }\!\!\}\!\!\text{ }$, which could be represented as $\text{ }\!\!\{\!\!\text{ }{\tilde{x}^{N-1}},{\tilde{x}^{N-2}},\cdots ,{\tilde{x}^{0}}\text{ }\!\!\}\!\!\text{ }$, where $\tilde{x}^{0} = {x}^{0}$ and $\tilde{x}^{s} = ({{x}^{s-1}}){{\uparrow }^{r}}$. \textcolor{black}{The ${\uparrow }^{r}$ represents the upsampling process}. Then, by combining the noise and the blurred version linearly, we could get the multi-scale diffusion process over $(s,t)$:
\begin{equation}
x_{t}^{s}=\sqrt{{{{\bar{\alpha }}}_{t}}}[\gamma _{t}^{s}{{\tilde{x}}^{s}}+(1-\gamma _{t}^{s}){{x}^{s}}]+\sqrt{1-{{{\bar{\alpha }}}_{t}}}\epsilon
\end{equation}
where the $\gamma _{t}^{s}$ is the mixing coefficient,which increases monotonically from 0 to 1. 

\begin{algorithm}
\caption{Training Procedure of SinDDM}
\label{alg1}
\begin{algorithmic}[1]
    \Repeat
    \State  $x^s \sim q(x^s)$
    \State  $t \sim Uniform(\{1,\dots,T\})$
    \State  $s \sim Uniform(\{1,\dots,N\})$ 
    \State  $\boldsymbol{\epsilon} \sim \mathcal{N}(\mathbf{0}, \mathbf{\mathit{I}})$ 
    \State  Update model ${\epsilon }_{\theta }$ by taking gradient descent step on:
    \State  $\nabla_\theta \left \|\boldsymbol{\epsilon} - \boldsymbol{\epsilon}_\theta(\sqrt{{{{\bar{\alpha }}}_{t}}}[\gamma _{t}^{s}{{\tilde{x}}^{s}}+(1-\gamma _{t}^{s}){{x}^{s}}]+\sqrt{1-{{{\bar{\alpha }}}_{t}}}\epsilon \right \|^2$ 
    \Until{converged}
\end{algorithmic}
\end{algorithm}

In the training process, a single fully convolutional model is trained to predict ${x}_{0}^{s}$ from the ${x}_{T}^{s}$. Because the noisy image is generated by adding the fixed noise according to the noise schedule, which is known, the goal of SinDDM is to gradually predict the noise $\epsilon$ added in the forward phase. First, we can model the recursive process of Markov chain in the single step of the diffusion process:
\begin{equation}
{{p}_{\theta }}(x_{t-1}^{s}|x_{t}^{s})=\mathcal{N}(x_{t-1}^{s};\mu _{\theta }^{s}(x_{t}^{s},t),\beta _{t}^{s}\textit{I})
\end{equation}
where the $\mu _{\theta }(x_{t}^{s},t)$ could be calculated as follows:
\begin{equation}
\mu _{\theta }(x_{t}^{s},t)=\frac{1}{\sqrt{{{\alpha }_{t}}}}(x_{t}^{s}-\frac{1-{{\alpha }_{t}}}{\sqrt{1-{{{\bar{\alpha }}}_{t}}}}\epsilon (x_{t}^{s},t)).
\end{equation}

In the optimization process, the KL divergence is optimized to make the predicted distribution ${{p}_{\theta }}(x_{t-1}^{s}|x_{t}^{s})$ closer to the real distribution
${{q}_{\theta }}(x_{t-1}^{s}|x_{t}^{s},x_{0}^{s})$. Thus, the optimization objective $\theta$ can be translated into calculating the Kullback-Leibler divergence $D_{KL}$ of the above two distributions:
\begin{equation}
\theta =\underset{\theta }{\mathop{\arg \min }}\,{{D}_{KL}}({{q}_{\theta }}(x_{t-1}^{s}|x_{t}^{s},x_{0}^{s}) \parallel {{p}_{\theta }}(x_{t-1}^{s}|x_{t}^{s})).
\end{equation}

Since the above distributions have the same variance, the Kullback-Leibler divergence can be simplified:
\begin{equation}
D_{KL}=\frac{1}{2\sum _{t}^{2}}{{\left\| \mu _{t}(x_{t}^{s},x_{0}^{s})-\mu _{\theta }(x_{t}^{s},t) \right\|}^{2}}+C
\end{equation}
where the $C$ is a constant and the $\sum _{t}^{2}$ is the covariance matrix of multivariate Gaussian distribution. Howerver, it is difficult to directly make the model predict $\mu _{\theta }$, because the distribution of the  $\mu _{\theta }$ is relatively uncertain and has a wide range of values, so the above prediction of the mean can be converted into a prediction of $\epsilon_{\theta}$.Then, through reparameterization techniques \cite{ho2020denoising}, the following simple loss function can be obtained:
\begin{equation}
L_{t-1}^{s}={{\mathbb{E}}_{x^{s}_{t},\epsilon }}\left[ {{\left\| \epsilon -{{\epsilon }_{\theta }}(\sqrt{{{{\bar{\alpha }}}_{t}}}{{x}^{s}_{t}}+\sqrt{1-{{{\bar{\alpha }}}_{t}}}\epsilon ,t,s) \right\|}_{1}} \right]
\end{equation}
where the ${\epsilon }_{\theta }$ is prediction of the fully convolutional model. The overall training procedure is shown in Algorithm \ref{alg1}.	

\textcolor{black}{With regard to} the diffusion process, the sampling via the reverse multi-scale diffusion starts from pure noise at the coarsest scale. In each scale, the SinDDM gradually removes the noise until reaching a clean image, which is then upsampled and combined with noise to start the process again in the next scale. The process will repeat until the scale $s=0$ comes to $s=N-1$. When $s=0$, we follow the standard approach, which starts with random noise at $t = T$ and gradually removing noise until a clean sample is obtained at $t = 0$. When $s>0$, we use $\sigma$ except when applying text-guidance in which case we also use the DDPM scheduler. As shown in Algorithm \ref{alg2}, the proposed model is conditioned on both the time step $t$ and the scale $s$, which is proved to improve generation quality and training time compared to a separate diffusion model for each scale.

\begin{figure}[!ht]\color{black}
\centering
\includegraphics[width =0.5\textwidth]{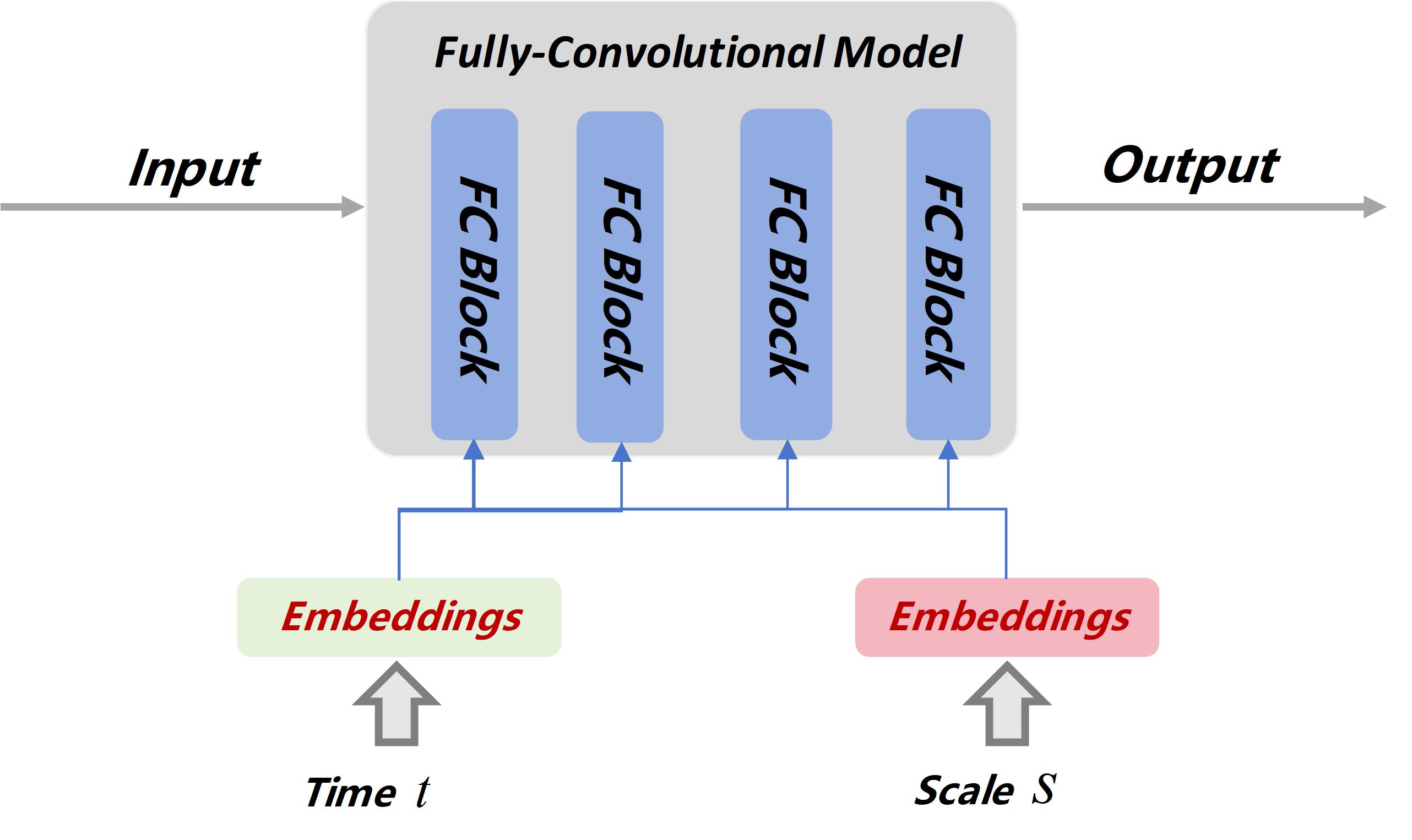}
\caption{The architecture of the proposed DDPM model. The fully-convolutional model is based on four blocks and uses the time step $t$ and scale $s$ as the condition to train.}
\label{structure}
\end{figure}

\begin{figure}[!ht] \color{black}
\centering
\includegraphics[width =0.5\textwidth]{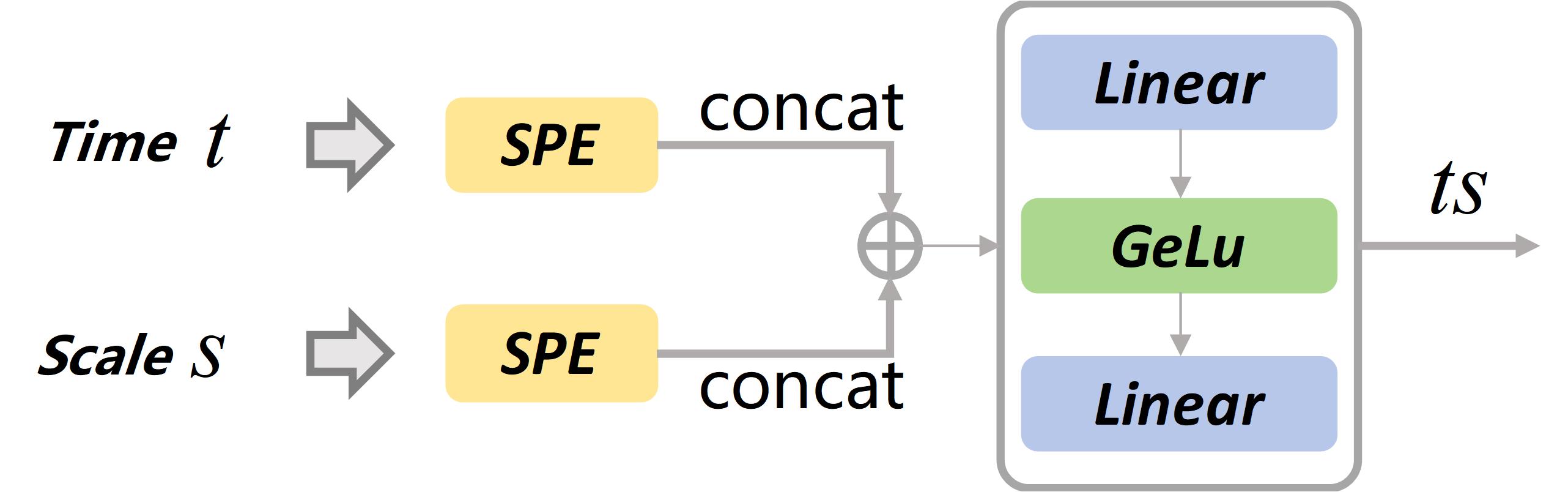}
\caption{The architecture of the embedding model of the time-step $t$ and scale $s$.}
\label{embedding}
\end{figure}

As shown in Fig. \ref{structure}, we use a fully-convolutional model with four blocks to achieve the process of noise prediction. As opposed to conventional DDPMs, the proposed model uses only convolutions and GeLU nonlinearities, without any self-attention or downsampling/upsampling operations. At the same time, we use the time step $t$ and scale $s$ as the condition to train the DDPM. Similar to the conventional DDPMs, which only embed time $t$ as a condition, we infuse the $t$ and $s$ to the model using a joint embedding. The time-step $t$ and scale $s$ first pass through an embedding block, in which they go through sinusoidal positional embedding (SPE) and then be concatenated and passed through two fully-connected layers with GeLU activation to yield a time-scale embedding vector $ts$. The SPE is derived from transformer model \cite{9716741}, which is a technique in which the position of each element in a sequence is embedded into a model. The main idea is to assign a unique code to each location by mapping each location to a fixed sinusoidal function in a high-dimensional space. Taking the SPE of $t$ as an example, the calculation formula is as follows:
\begin{equation}
\begin{split}
    & SPE(t,2i)=\sin (t/{{10000}^{2i/d}}) \\ 
    & SPE(t,2i+1)=\cos (t/{{10000}^{2i/d}}) \\ 
\end{split}
\end{equation}
where the $d$ is the embedding dimension of the $s$ after SPE and the $i$ represents the number of the dimension. Similarly, the scale parameter $s$ can also be obtained by the SPE and has the same size as the code value of $t$.The embedding procedure is shown in the Fig. \ref{embedding}.

\begin{figure}[!ht] \color{black}
\centering
\includegraphics[width =0.5\textwidth]{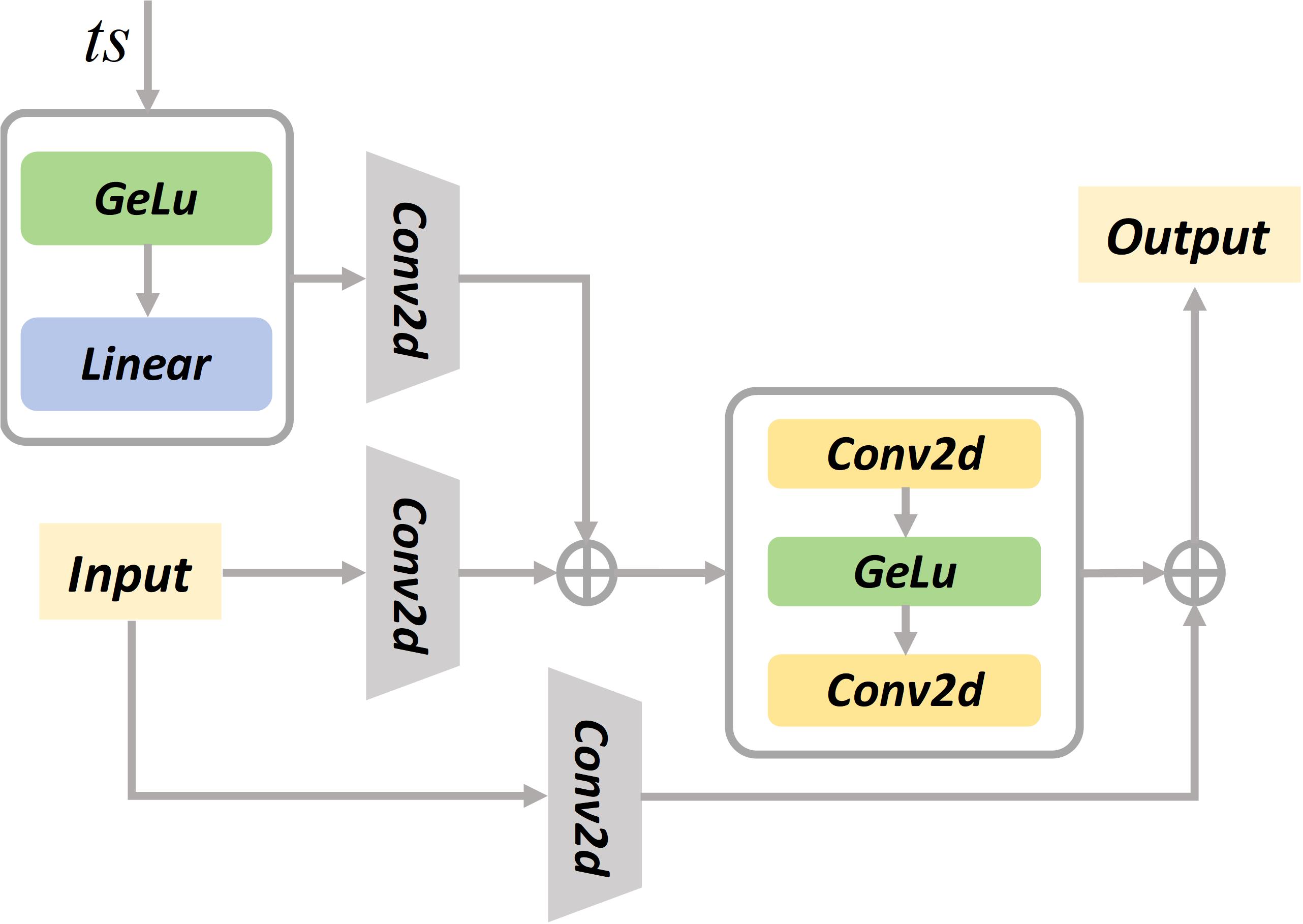}
\caption{The architecture of the block.}
\label{block}
\end{figure}

As for the fully-convolutional blocks, as shown in Fig. \ref{block}, the pipeline consists of 2D convolutions with a residual connection and GeLU function, and the input time-scale embedding $ts$ vector is passed through a GeLU activation and a fully-connected layer. In each training iteration we sample a batch of noisy images from the same randomly chosen scale $s$ but from several randomly chosen time steps $t$. We used the Adam optimizer to train 12,000 epochs during the training procedure.

\begin{algorithm}
\caption{Sampling Procedure of SinDDM}
\label{alg2}
\begin{algorithmic}[1]
\For{$s=0,1,2,\cdots, N-1$} 
\If{$s=0$}
\State $x_{T[0]}^{0}\sim \mathcal{N}(\boldsymbol{\mathit{0,I}})$
\EndIf
\For{$t = T[s],T[s-1],\cdots,0$}
\State $x_{t}^{s,mix}=\frac{x_{t}^{s}-\sqrt{1-{{{\bar{\alpha }}}_{t}}}{{\epsilon }_{\theta }}(x_{t}^{s},t,s)}{\sqrt{{{{\bar{\alpha }}}_{t}}}}$
\State $\hat{x}_{0}^{s}=\frac{x_{t}^{s,mix}-\gamma _{t}^{s}{{{\tilde{x}}}^{s}}}{1-\gamma _{t}^{s}}$
\State $x_{t-1}^{s,mix}=\gamma _{t-1}^{s}{{\tilde{x}}^{s}}+(1-\gamma _{t-1}^{s})\hat{x}_{0}^{s}$
\State $\epsilon \sim \mathcal{N}(\boldsymbol{\mathit{0,I}})$
\State $x_{t-1}^{s}=\sqrt{{{{\bar{\alpha }}}_{t}}-1}x_{t-1}^{s,mix} +$ 
\State $\sqrt{1-{{{\bar{\alpha }}}_{t-1}}-{{(\sigma _{t}^{s})}^{2}}}\frac{x_{t}^{s}-\sqrt{{{{\bar{\alpha }}}_{t}}}x_{t}^{s,mix}}{\sqrt{1-{{{\bar{\alpha }}}_{t}}}}+\sigma _{t}^{s}\epsilon$
\EndFor
\If{$s< N-1$}
\State ${{\hat{x}}^{s+1}}=\hat{x}_{0}^{s}{{\uparrow }^{r}}$
\State $\epsilon \sim \mathcal{N}(\boldsymbol{\mathit{0,I}})$
\State $x_{T[s+1]}^{s+1}=\sqrt{{{{\bar{\alpha }}}_{T[s+1]}}}{{\tilde{x}}^{s+1}}+\sqrt{1-{{{\bar{\alpha }}}_{T[s+1]}}}\epsilon$
\EndIf
\EndFor
\end{algorithmic}
\end{algorithm}

\subsection{The Prompt-guided Fine-tuning Process}	

\begin{figure*}[!ht] \color{black}
\centering
\includegraphics[width = 1.0\textwidth]{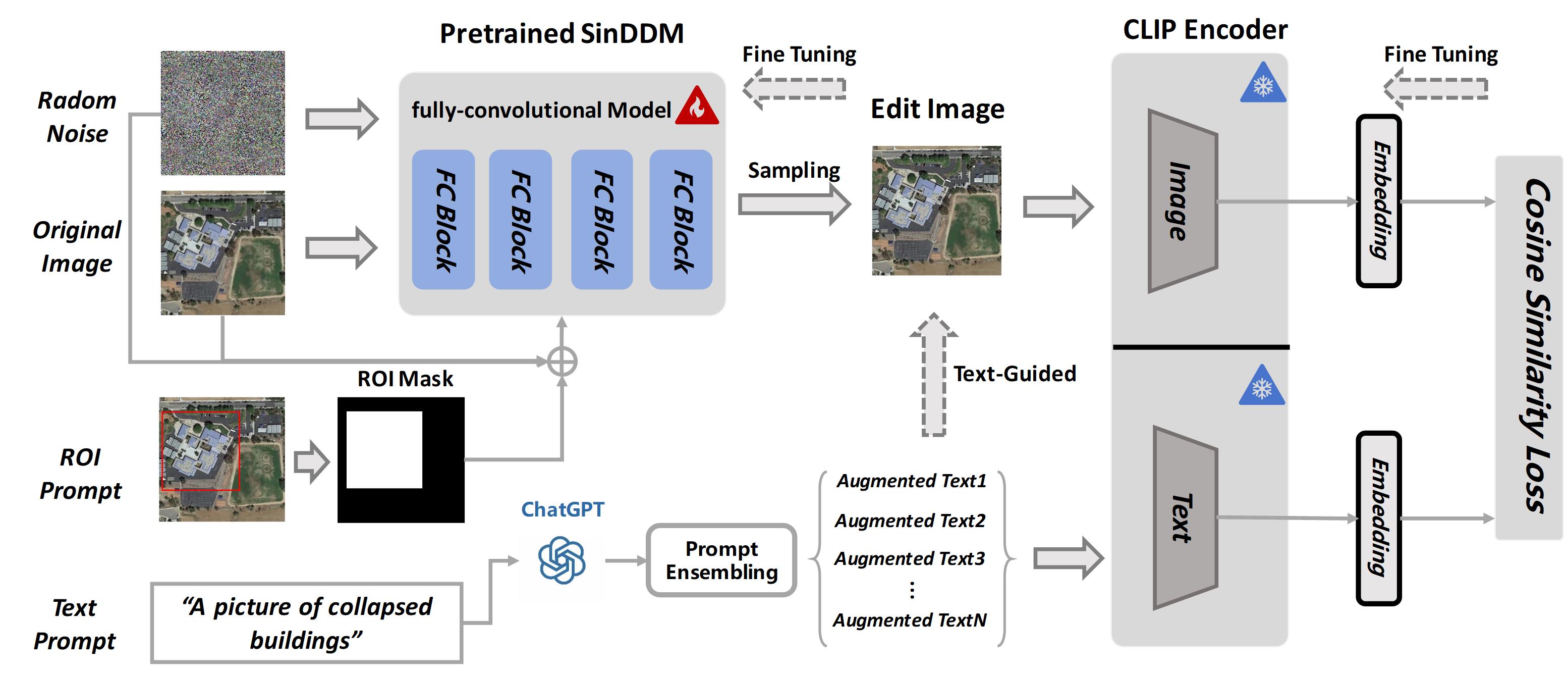}
\caption{The prompt-guided fine-tuning process of SinDDM. After the training procedure, the encoder of the SinDDM is frozen and its decoder is fine-tuned through prompt's guidance. The ROI prompt is added to the original image to mask the areas that do not need editing, while the text prompts would be augmented with PE. Then, we use the frozen image encoder and text encoder of the CLIP to generate corresponding embeddings of the edited image and text prompt, respectively. Finally, we use cosine similarity loss between the two above embeddings as the optimization objective to fine tune the decoder of the SinDDM.}
\label{finetune}
\end{figure*}

The core objective of image editing is to incorporate new semantic information, typically in the form of ROIs or text, known as prompts. Various methods exist for embedding semantic information from prompts into DDPMs \cite{huang2024}. In our proposed approach, we leverage CLIP to extract text embeddings from the prompts and apply CLIP loss as supervision to fine-tune the sampling process of SinDDM, as depicted in Fig. \ref{finetune}.

Specifically, the trained SinDDM is fine-tuned using different types of prompt guidance. The ROI prompt is used to mask regions of the original image that do not require modification, while text prompts are refined with PE. Given its alignment across hundreds of millions of image-text pairs, CLIP contains rich multi-modal representational information, making it an optimal choice for incorporating text-based semantic information.

To use CLIP as the guidance, first, the current generated image $\hat{x}_{0}^{s}$ and the text prompts provided by the users are sent to the image encoder ${{f}_{I}}(\cdot )$ and text encoder ${{f}_{T}}(\cdot )$ of the CLIP model, respectively. Then, we use the cosine similarity loss function to measure the discrepancy between the above two embeddings and regard it as an optimization target in the process of fine-tuning. Therefore, the CLIP loss $L_{CLIP}$ can be depicted as follows:
\begin{equation}
{{L}_{CLIP}}=-\frac{{{f}_{I}}(\hat{x}_{0}^{s})\cdot {{f}_{T}}(\text{text})}{\left| {{f}_{I}}(\hat{x}_{0}^{s}) \right|\cdot \left| {{f}_{T}}(\text{text}) \right|}.
\end{equation}

The prompts provided by the users can be divided into two types: when they are only text prompts, the above CLIP-guidance stops at $s = N-1$ to produce a smoother edited image. However, if the prompts include both the ROI limiting the editing area and the corresponding text, the affected regions would be constrained the spatial extent by zeroing out all gradients outside the ROI mask. This mask is calculated in the first step CLIP is applied, and is kept fixed for all remaining time steps and scales (it is upsampled when going up the scales of the pyramid). We define the mask generated by ROI as $m^s$, then during the fine tuning procedure, we could update the $\hat{x}^{s}_{0}$ via the following formula:
\begin{equation} \color{black}
\hat{x}_{0}^{s+1} = \eta \delta {{m}^{s}}\odot \nabla {{L}_{CLIP}}+(1-{{m}^{s}})\odot \hat{x}_{0}^{s}
\end{equation}
where the $\eta \in [0,1]$ is an strength parameter that controls the intensity of the CLIP
guidance and the $\delta =\left\| \hat{x}_{0}^{s}\odot m \right\|/\left\| \nabla {{L}_{CLIP}}\odot m \right\|$. Due to the strong prior of the proposed denoiser, which is overfitted to the statistics of the training image, the simple update steps tend to be ineffective. In order to avoid the resolution of the original CLIP gradient at each denoiser step, the update rule of $\hat{x}^{s}_{0}$ will be transformed like follows with a momentum:
\begin{equation} \color{black}
\begin{split}
    \hat{x}_{0}^{s+1} = \eta \delta {{m}^{s}}\odot \nabla {{L}_{CLIP}} + &(1-{{m}^{s}})\odot \\
    &(\lambda \hat{x}_{0}^{s}+(1-\lambda )\hat{x}_{0,prev}^{s})
\end{split}
\end{equation}
where the $\hat{x}_{0,prev}^{s}$ is the $\hat{x}^{s}_{0}$ from the previous time-step and the $\eta$ is a momentum parameter which is set to 0.05. 

Meanwhile, the original image ${x}_{0}^{s}$ is added with the same noise and blended with the masked image to continue the diffusion process, which will improve the smooth transition between the edge of the edited content in the ROI region and the original image. It is worth noting that, since the CLIP model's image and text encoder is only used to generate the embeddings during the fine-tuning process, their parameters are also frozen and do not participate in back propagation.

Since in image editing, in addition to using text as guidance, the original region prompt in the image is also crucial, especially when the texts could not accurately describe the edited target. At this point, our proposed method use ROI to choose a specific area of the image and copy the content of the specific area at the specified location to achieve ROI-prompted image editing. In this way, the user can choose to replicate the targets to a specific area, which is very useful for some repetitive targets, such as tightly packed vehicles, oil storage tanks laid out in port terminals, and wind turbines and photovoltaic panels arranged on various scenes. To be more specific, we use $x_{target}^{s}$ to be an image containing the desired contents within the target ROIs, while the $m^s$ be a binary mask indicating the ROIs, both down-sampled to scale $s$. Then we use an simple $L^2$ loss to achieve the ROI guidance:
\begin{equation}
\begin{split}
    {{L}_{ROI}}={{\left\|  (\hat{x}_{0}^{s}-\hat{x}_{target}^{s}) \odot{{m}^{s}} \right\|}^{2}}
\end{split}. 
\end{equation}

In the sampling procedure, we update the $\hat{x}_{0}^{s}$ by taking a gradient step on this loss, while mix $\hat{x}_{target}^{s}$ and $\hat{x}_{0}^{s}$ with the step size $\eta$. Namely:
\begin{equation} \color{black}
\begin{split}
    \hat{x}_{0}^{s+1} =  ((1-\eta )\hat{x}_{0}^{s}&+ \eta \cdot x_{target}^{s})\odot {{m}^{s}} \\
    &+(1-{{m}^{s}})\odot \hat{x}_{0}^{s}
\end{split}
\end{equation}
where the $\eta$ determines the strength of the effect. We use this guidance in all scales except for the finest one.

\subsection{Prompt Ensembling}

\begin{figure}[!ht]
\centering
\includegraphics[width = 0.4\textwidth]{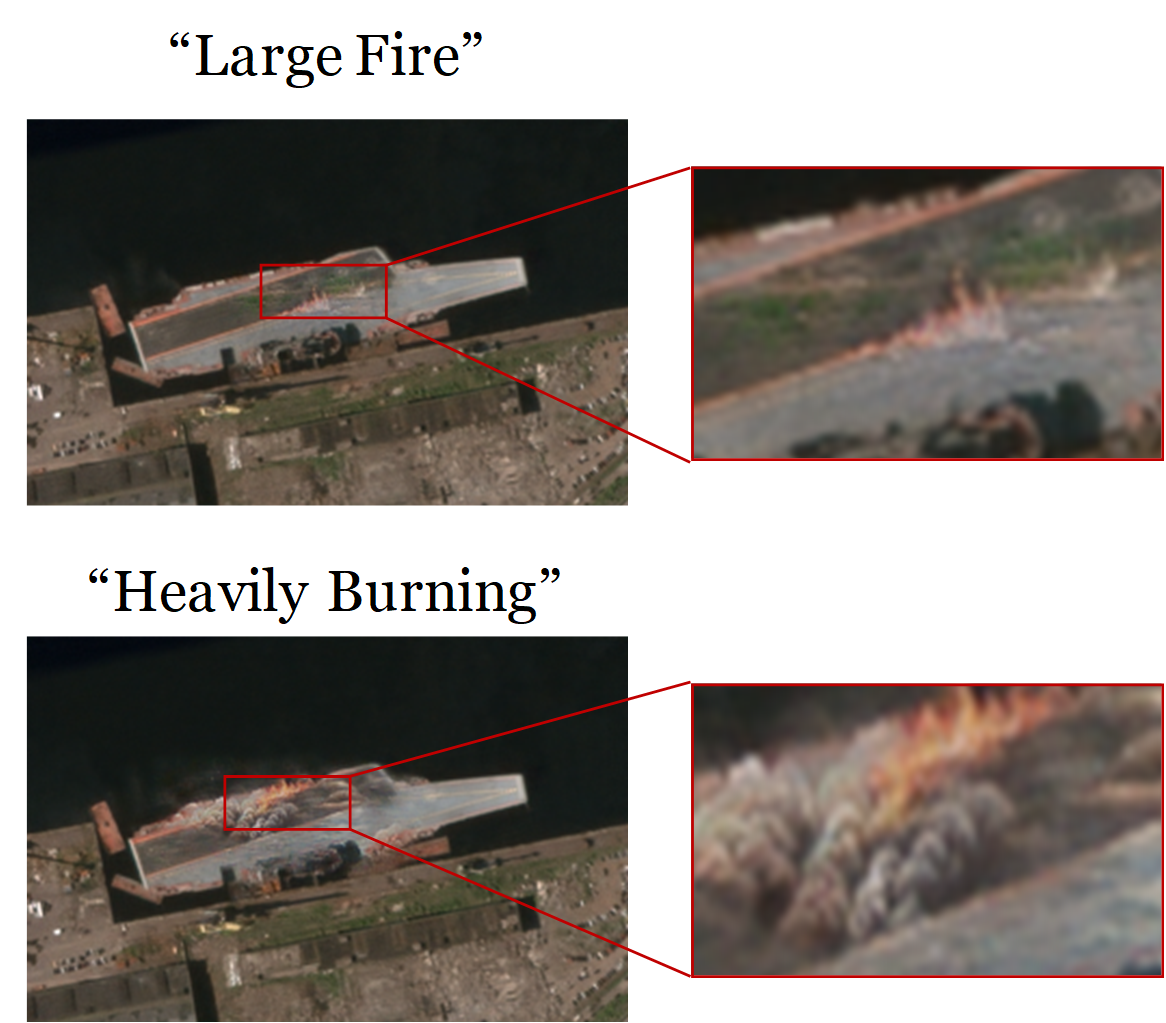}
\caption{The generated images with different text prompts. When the prompt is ``Large Fire", the edited image area will have an unreasonable result of grass on the deck. However, when the prompt is ``heavily burning", the results will much more logical.}
\label{before-PE}
\end{figure}

Recently, prompt based fine-tuning is widely used in the application of multiple foundation models \cite{Khattak_2023_CVPR, Liu_2023_CVPR, Park_2024_CVPR}. Unlike conventional fine-tuning approaches, the prompt-based paradigm harmoniously unifies the trained DDPMs and downstream tasks within the same framework. Thus, choosing the right prompt is critical to the effectiveness of downstream tasks. For the text-guided image editing method, the text prompts provided by the users have a great influence on the result of image editing, especially if the CLIP itself is potentially biased. For example, when the user uses ``Large Fire" as text prompt to generate an image depicting a ship on fire, since ``Fire” is often associated with grassland in the semantic information of CLIP, the edited image area will have an unreasonable result of grass on board. Nevertheless, when the user used ``heavily burning" as the prompt, the results will much closer to the authentic situations. Fig. \ref{before-PE} depicts the influence of the different text prompts.

\begin{figure}[!ht]
\centering
\includegraphics[width = 0.4\textwidth]{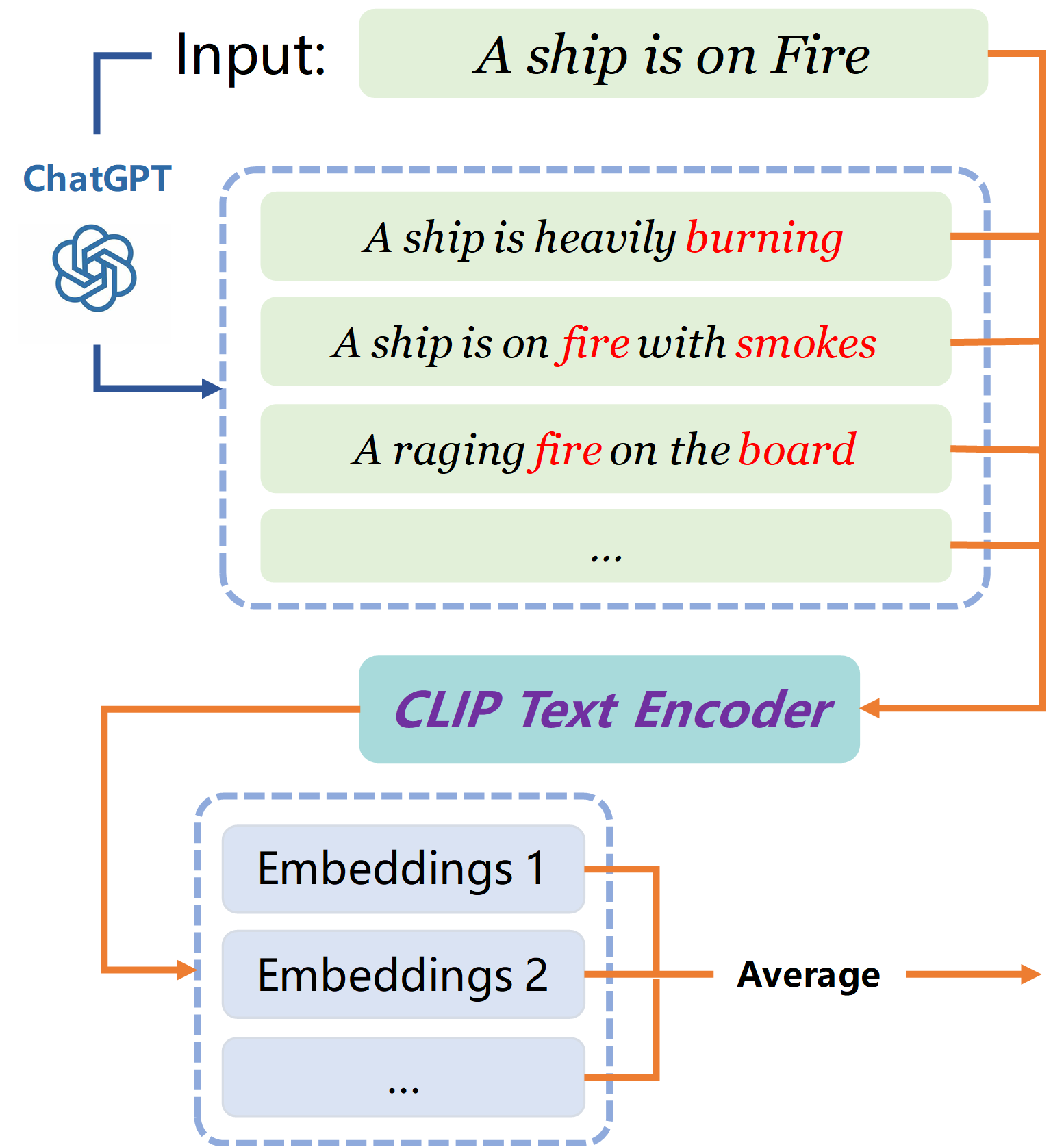}
\caption{The overall structure of PE. The text prompt provided by the user is converted into a set of various synonymous texts via GPT, and then the text set is embedded and averaged by the CLIP text encoder as a guide for image editing.}
\label{PE}
\end{figure}

The reason for the aforementioned unsatisfactory generation results lies in the fact that, within CLIP, text semantics and image semantics exhibit a one-to-many correspondence. This makes it challenging to find an ideal image embedding using a single text embedding. Recent studies have demonstrated that employing multiple prompts can further enhance the performance of prompt learning, a technique known as multi-prompt learning. Among the most straightforward and effective approaches, PE mitigates the generation of irrelevant content by combining multiple text prompts with the same semantics expressed in different forms. Consequently, we employ PE in our proposed method as a pre-processing strategy to produce robust text guidance. Additionally, for diversity, we leverage ChatGPT as a generation tool to quickly produce high-quality prompts. The overall structure is illustrated in Fig.\ref{PE}.

Utilizing LLMs such as ChatGPT, the original text is leveraged to generate various semantically similar text prompts. These augmented text prompts enhance the information provided by the user and transform it into multiple forms. For instance, when the user's input is ``A ship is on fire," we use the following command to instruct ChatGPT to generate corresponding augmented text prompts:

\textit{``I need you act as a text prompt generator. I will give you a text prompt that you need to use common sense to translate into a total of five prompts with different descriptions but similar meanings. Each prompt given meets the same syntax format as the original text prompt and is concise and easy to understand. Here is an example: Given ``A ship is on fire", you need to take into account the actual scenario of the ship on fire and generate the approximate description ``A ship is burning". You only need to provide the generated text prompt, you do not need to explain why."}

Through the above instructions, we can use ChatGPT to generate numerous augmented text, for the text prompt ``A ship is on fire", the augmented texts could be ``A ship is burning", ``A ship is on fire with smokes", ``A raging fire on the board'' and so on. After that, we obtain the embeddings of the prompts separately from the text encoder of CLIP, and average them as the guidance of the image editing model.

\section{Experiments}
\subsection{\textcolor{black}{Experiment Settings}} 

\begin{figure*}[!ht]\color{black}
\centering
\includegraphics[width = 1.0\textwidth]{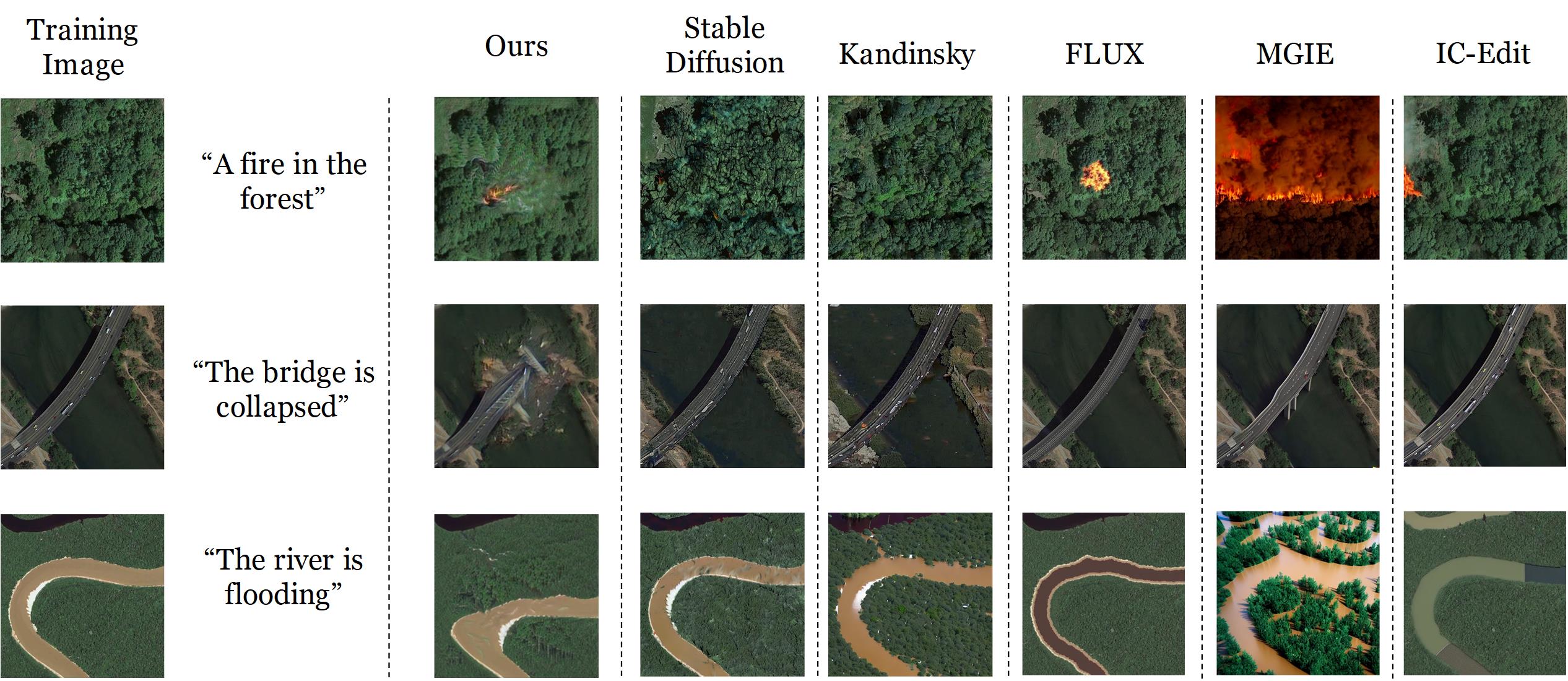}
\caption{Generated images of the our proposed method, the Stable Diffusion, Kandinsky, FLUX, MGIE and IC-Edit under three scenarios. The above images show the results of full image editing under three scenarios: ``A fire in the forest", ``The bridge is collapsed" and ``The river is flooding".}
\label{full_image}
\end{figure*}

We evaluate our text-guided RSI editing model through two editing scenarios. First, when comes to the scenario where the type of ground objects is homogeneous and the overall editing is required, we adopt the method of directly repainting the whole image through text guidance to edit the image. Secondly, in the scenario where the types of ground objects are complex and the editing of local areas needs to be refined, we edit it by combining the text guidance and ROI region mask. For the above two scenarios, we take two kinds of datasets: AID \cite{7907303} and HRSC2016-MS \cite{Chen2022MSSDet}. 

For the selection of VLMs, in order to introduce the semantic information of RSIs, we use the RemoteCLIP model \cite{remoteclip}, which is fine-tuned on remote sensing big data. \textcolor{black}{To validate the quality of the edited images quantificationally, we also use CLIP score \cite{DBLP:journals/corr/abs-2104-08718} and a subjective evaluation metric to compare the generation results between our image editing method and mainstream image editing methods such as Stable Diffusion \cite{rombach2021highresolution}, Kandinsky \cite{kandinsky2.1}, FLUX \footnote{https://huggingface.co/spaces/black-forest-labs/FLUX.1-Fill-dev}, MLLM-Guided Image Editing (MGIE) \cite{fu2024} and In-Context Edit (IC-Edit) \cite{zhang2025incontexteditenablinginstructional}. }\textcolor{black}{CLIP score is calculated by using the CLIP encoder to acquire the features of the edited image and text prompts respectively and computing their cosine similarity. Existing researches \cite{DBLP:journals/corr/abs-2104-08718}have shown that CLIPScore achieves the highest correlation with human judgments, the calculation formula is as follows:}
\begin{equation}\color{black} \color{black}
c =\omega \cdot \max (\langle {{e}_{i}},{{e}_{t}} \rangle,0),
\end{equation}
\textcolor{black}{where the $\omega$ is set to 1.0 and ${e}_{i},{e}_{t}$ refers to the visual CLIP embedding and textual CLIP embedding respectively. The $c$ is the CLIP score and the operator $\langle \cdot, \cdot \rangle$ represents the cosine similarity.} 

\begin{figure*}[!ht]
\centering
\includegraphics[width = 1.0\textwidth]{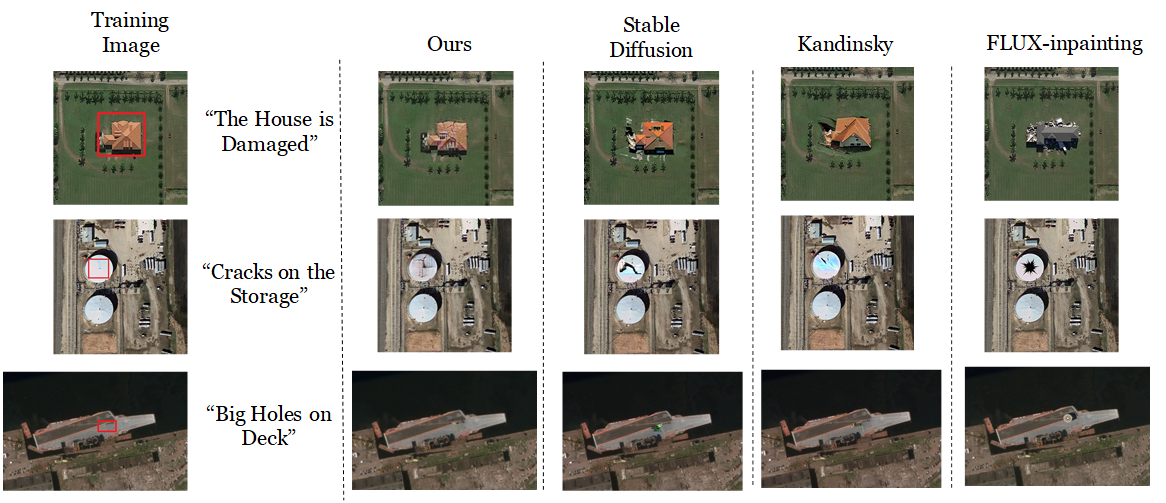}
\caption{Generated images of the our proposed method, the Stable Diffusion, Kandinsky and FLUX (inpainting) under three scenarios. The above images show the results of local image editing under different scenarios based on the masks: ``The house is damaged", ``Cracks on the storage" and ``Big holes on deck".}
\label{impainting}
\end{figure*}

For the subjective score, we made a suitable questionnaire, using image fidelity and edited quality as the evaluation focus, and asked the experts to give a score from 1 to 5 for each image in the above two aspects. Then, we calculate the comprehensive subjective score by averaging the scores. When conducting a survey, we provide the original image, the requirements for image editing, and the generated images that been edited by different methods, where the edited images are randomly scrambled and not ordered according to the methods. For each edited image, the question we asked the respondents was: 

\textit{``Please give the scores of the fidelity and the degree of the satisfaction of the needs according to the original image provided and the corresponding editing requirements. The scores are integers from 1 to 5."}

We collected a total of about 200 survey results, about 40\% of the interviewee are professional RSI interpreters, about 60\% are outside the field of respondents, the knowledge of RSIs is limited to a preliminary understanding. Therefore, the obtained questionnaire can not only assess the satisfaction of the editing needs with the help of the professional knowledge of remote sensing interpreters, but also evaluate the degree of its fidelity with a large number of non-professionals.

\subsection{\textcolor{black}{Experiment Results}}
First of all, we test the full image editing performance of each model. There are three scenarios: ``A fire in the forest", ``The bridge is collapsed" and ``The river is flooding". The generated images are depicted in the Fig. \ref{full_image}. Intuitively, our generation model is much more complete with the details of the original image beyond the obvious editorial content. When asked to generate a forest on fire, the image produced by our model had a clear flame present, while other models could not. The same goes for the editorial content of the ``collapsed bridge", it can be seen that the images generated by ours faithfully recreate the collapse scene while others just distort the bridge in the image. \textcolor{black}{With regard to} the flooding river, in the images we generated, the river channel is not only biased but also has an impact on the surrounding forest, which is in good agreement with the actual situation.

Different from full image editing, local image editing guided by mask is often more concerned with the degree of refinement and the fusion of mask edges with the original image. To do this, we design three scenarios of image editing tasks focusing on the targets: ``The house is damaged", ``Cracks on the storage" and ``Big holes on deck". The generated images are depicted in the Fig.\ref{impainting}. Compared to other editing methods, our proposed method also achieved the best results. When generate the image of storage with cracks, our method produces more realistic cracks, which can even simulate shadows and rust caused by the cracks. When generate the damaged house, our proposed method better preserves the outline of the target house, while stable diffusion and Kandinsky significantly change the style of target house. Similarly, when generating ship with holes on deck, our results are also more in line with human visual habits. 

\textcolor{black}{In contrast, the images generated by Stable Diffusion, Kandinsky, FLUX, MGIE and IC-Edit display a pronounced phenomenon of semantic confusion and inconsistency. For example, when large cracks are introduced to storage tanks, all generated images—except those produced by our proposed method—exhibit unreasonable colors. Although the images generated by FLUX fully demonstrate the concept of damage, the distortion is severe. At the same time, due to MGIE's over-reliance on LLM and pre-trained generative models for image editing, it cannot maintain the semantics of the original image, causing the edited image to be out of the remote sensing perspective. Furthermore, when generating an image of a damaged house, Kandinsky presents an edited depiction from a non-remote sensing perspective. Similarly, when creating large holes on a deck, all aforementioned methods produce irrelevant content in their outputs. These generative flaws can be attributed to several primary factors mentioned in this paper. Firstly, conventional models are trained on large benchmark datasets of natural images, which exhibit significant domain discrepancies compared to remote sensing images. Consequently, these generative models with low generalization struggle to preserve consistency in content and details between the original and edited images, leading to illogical results during image generation. Additionally, these models have been predominantly pre-trained on street-view imagery, resulting in a significant semantic gap between them and the bird's-eye view RSIs targeted for editing. Lastly, semantic mismatches between textual descriptions and visual representations further exacerbate the confusion observed in the editing results.}


\begin{table}[!ht] \color{black}
\centering
\caption{Quantitative evaluation results of comparative methods. The highest score is highlighted in bold. SD and Kan represent Stable Diffusion and Kandinsky, respectively.}
\begin{tabular*}{\hsize} {@{}@{\extracolsep{\fill}}p{0.8cm}cccccccccc@{}} 
    \hline
    \textbf{} & \textbf{} & Ours & SD & Kan & FLUX & MGIE & IC-Edit\\ \hline
    \multirow{6}{*}{\makecell{CLIP \\ Score (\%)}} & Forest & \textbf{21.38} & 19.94 & 17.24 &20.46  &21.07 &20.76\\ 
    \textbf{} & Bridge & \textbf{22.64} & 21.08 & 21.26 &20.21  &20.38 &20.37\\ 
    \textbf{} & River & \textbf{20.11} & 19.38 & 19.16 &19.07  &19.98 &19.12\\ 
    \textbf{} & House & \textbf{22.76} & 21.98 & 21.55 &21.65 & / & /\\ 
    \textbf{} & Storage & \textbf{17.98} & 15.48 & 15.94 &16.07 & / & /\\ 
    \textbf{} & Ship & \textbf{14.68} & 10.83 & 11.55 &12.50 & / & /\\          \hline
    \multirow{6}{*}{\makecell{Subjective \\ Score}} & Forest & \textbf{4.50} & 4.15 & 3.76 & 3.89  &3.27 & 3.42 \\   
    \textbf{} & Bridge & \textbf{4.56} & 4.03 & 4.11 &4.04  & 4.45 & 4.09 \\ 
    \textbf{} & River & 4.25 & 3.93 & \textbf{4.41} &4.18  &4.06  & 4.12 \\ 
    \textbf{} & House & \textbf{4.65} & 4.25 & 4.13 &4.27 & / & / \\ 
    \textbf{} & Storage & \textbf{4.59} & 4.27 & 4.06 &4.38 & / & / \\ 
    \textbf{} & Ship & \textbf{4.44} & 3.94 & 4.23 &3.72  & / & / \\ \hline
\end{tabular*}
\label{metrics}
\end{table}

\begin{figure}[!ht]
\color{black}
\centering
\includegraphics[width = 0.5\textwidth]{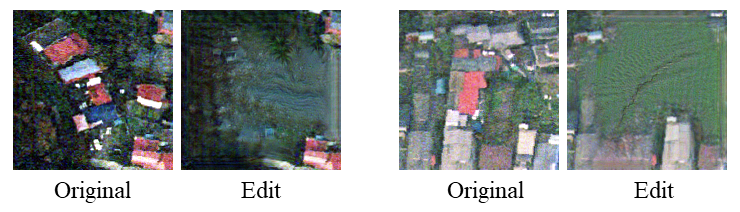}
\caption{RSIs of buildings before the tsunami and corresponding post-tsunami images generated in prompt ``a picture of ground after tsunami". Here are some examples.}
\label{building_edit}
\end{figure}

The Table \ref{metrics} shows the quantitative evaluation metrics of edited images generated by different models under the various scenarios. Firstly, as to the CLIP score, our proposed model achieves optimal results in all scenarios. This shows that the images generated by our model are more consistent with the content of the edited text. In addition to objective evaluation metrics, subjective evaluation is equally important for evaluating image quality. After expert testing, the image generated by our proposed model achieved the highest scores in most scenarios, which shows that they're more consistent with human vision and more logical. What we have to admit is that due to length limitations, more experimental results will not be presented in the paper, but they are available at \href{https://github.com/HIT-PhilipHan/remote_sensing_image_editing}{\textcolor{black}{https://github.com/HIT-PhilipHan/remote\_sensing\_image\_editing}.} 

\begin{table}[!ht]\color{black}
\centering
\caption{The experimental results of building damage estimation with different training set.}
\begin{tabular*}{\hsize}{@{}@{\extracolsep{\fill}}ccccccccccccc@{}}
    \hline
    ~ & 5 pairs & 10 pairs & 30 pairs & 50 pairs \\ \hline
    Only Observed Samples & 68.74 & 71.82 & 81.66 & 89.42  \\ \hline
    With Generated Samples & 70.92 & 73.39 & 85.14 & 91.07 \\ \hline
\end{tabular*}
\label{building_estimation}
\end{table}

\begin{figure}[!ht]\color{black}
\centering
\includegraphics[width =0.5\textwidth]{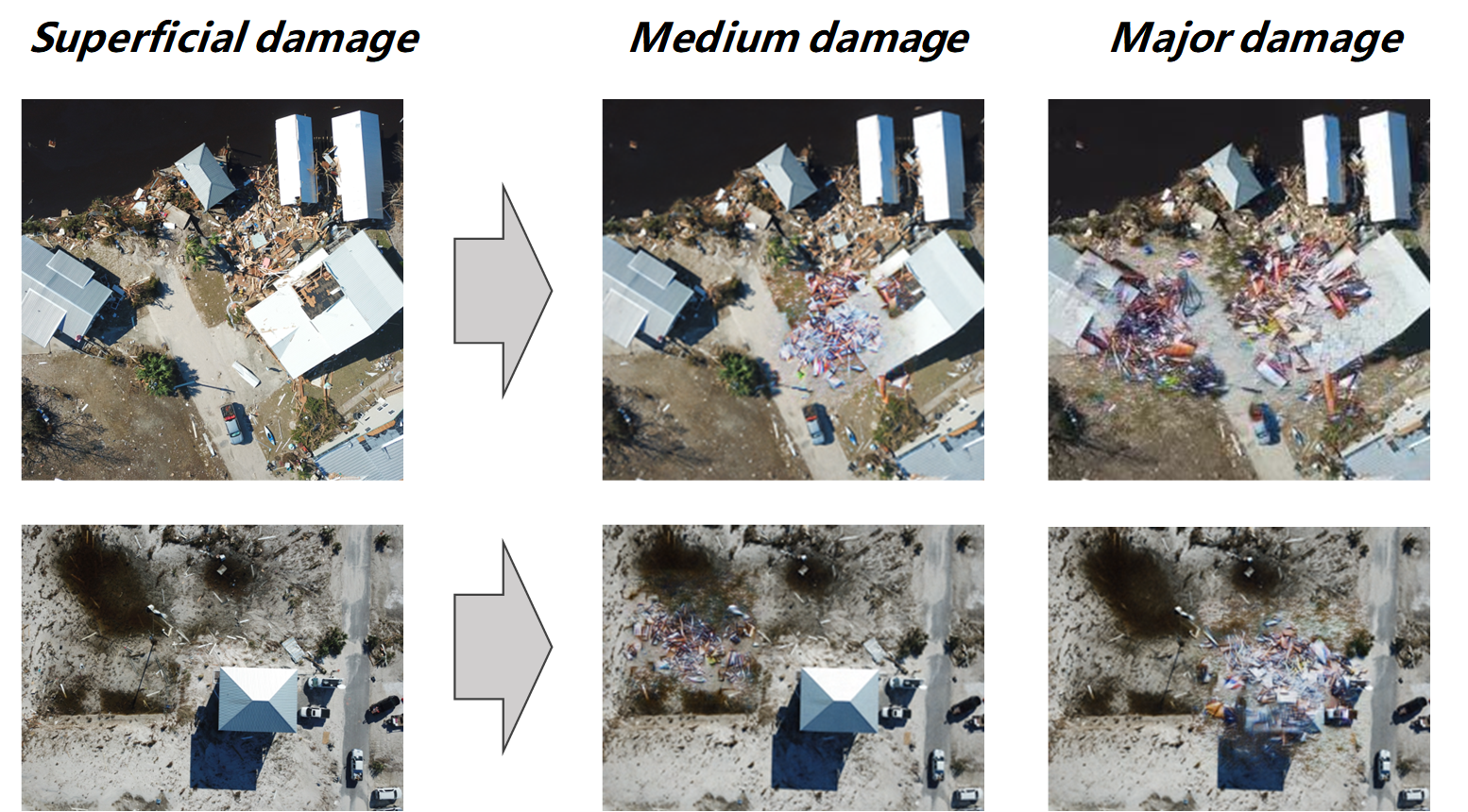}
\caption{RSIs of building scenes with different damage levels generated in prompt ``A mess of house debris". Here are some examples.}
\label{building_damge}
\end{figure}

\begin{table}[!ht]\color{black}
\centering
\caption{The class probabilities of building damage estimation of original and edited images. The asterisk ($*$) indicates the data with the highest class probability.}
\begin{tabular*}{\hsize}{@{}@{\extracolsep{\fill}}ccccccccccccc@{}}
\hline
    ~ & ~ & Image1 & Image2 \\ \hline
    ~ & Superficial & $0.3496^*$  & $0.3435^*$ \\ 
    Original & Medium & 0.3137 & 0.3226 \\ 
    ~ & Major & 0.3367 & 0.3339 \\ \hline
    ~ & Superficial & 0.3216 & 0.3251 \\ 
    Edited1 (Small ROI) & Medium & $0.3410^*$ &  $0.3551^*$ \\ 
    ~ & Major & 0.3374 & 0.3198 \\ \hline
    ~ & Superficial &0.3147  & 0.3258 \\ 
    Edited2 (Large ROI)& Medium &0.3369  & 0.3277 \\ 
    ~ & Major &$0.3484^*$  & $0.3465^*$ \\ \hline
\end{tabular*}
\label{building_damage_edit}
\end{table}

\begin{figure*}[!ht]
\centering
\includegraphics[width = 1.0\textwidth]{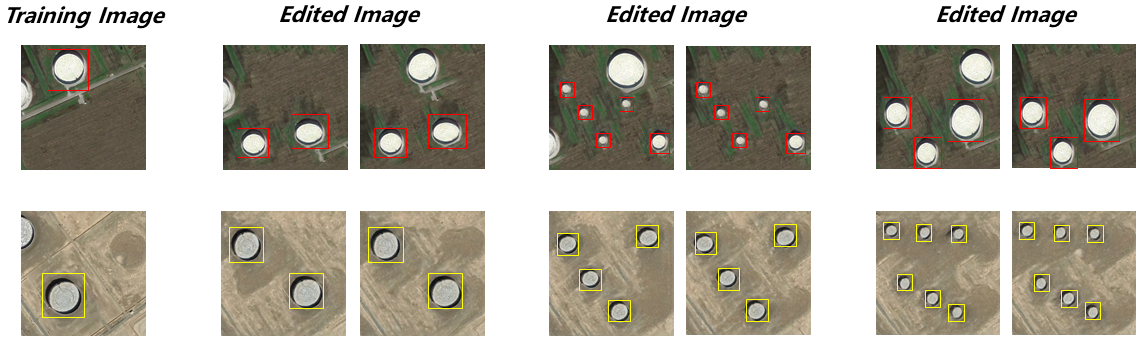}
\caption{The edited images of the oil storage tanks via the ROI-prompted image editing method.}
\label{oiltank}
\end{figure*}

\subsection{\textcolor{black}{Discussion}} \color{black}

To validate the proposed image editing method for supporting remote sensing tasks, we used the disaster change detection and assessment dataset \cite{7986759}, generating post-disaster images with the prompt ``a picture of ground after tsunami" to construct the training dataset, some samples are shown in Fig. \ref{building_edit}. Using a Siamese network based on ResNet18, we input pre- and post-disaster images to estimate building destruction. We combined generated data pairs with real observation data pairs and compared test set accuracy. Each experiment included 30 pairs of undamaged house images additionally. The results, shown in Table \ref{building_estimation}, demonstrate that our generated post-disaster images effectively supplement real observation images. When the experiment is set to 5 pairs, 10 pairs, 30 pairs and 50 pairs, the prediction accuracy of the model after adding the generated image pairs is higher than that using only the observation data, and the accuracy improvement is 2.45\%, 1.57\%, 3.48\% and 1.65\% respectively. 

\begin{figure}[!ht]
\centering
\includegraphics[width =0.5\textwidth]{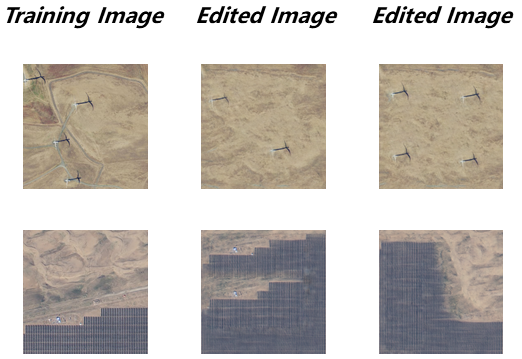}
\caption{More applications of the image editing method, for example, generating photovoltaic panels, wind turbine and so on at specified locations in the RSIs.}
\label{application}
\end{figure}

\textcolor{black}{In addition to measuring tsunami damage, we also conduct experiments on the RescueNet dataset \cite{Rahnemoonfar2023}, a high resolution UAV classification dataset for natural disaster damage assessment to evaluate the editing effect. Each image in the dataset is classified into one of three classes: superficial damage, medium damage, and major damage. The classification of an image is based on the extent of damage within the area covered by that image, encompassing both man-made and natural structures. If the image does not exhibit any damaged structures, it is classified as ``Superficial damage''. In the case where a few structures are damaged by the natural disaster, the image is labeled as ``Medium damage''. Finally, if the image contains at least one completely destroyed building or if approximately 50\% of the area is covered with debris, it is categorized as ``Major damage''.}

\textcolor{black}{To validate that the edited images can effectively support downstream tasks, we aim to ensure that a classification model trained on the dataset can accurately identify the degree of building damage depicted in the edited images. Accurate classification results would demonstrate that the edited images can serve as viable substitutes for real images. Consequently, we conducted image editing on pictures showing superficial damage, using the text prompt ``A mess of house debris'' and generated images representing medium and major damage by selecting ROIs of varying sizes. We modified two images depicting superficial damage in the original RescueNet dataset and simulated medium damage by incorporating house debris within a small-sized ROI. Similarly, we represented major damage by adding debris of houses within a large-sized ROI. The editing results are shown as Fig. \ref{building_damge}.}

\textcolor{black}{The classification model employed in our experiments is ResNet18, and the class probabilities for both the original and edited images are shown in Table \ref{building_damage_edit}. To be more specific, after editing the images of superficial damage in the original dataset using different ROI sizes, the model's predicted results shift from superficial damage to the medium damage. For instance, regard as the unedited Image1, a superficial damage image have the highest classification accuracy, with values of 34.96\%, 31.37\%, and 33.67\% for the three categories, respectively. When applying edits using a small-sized ROI and text prompt, the classification accuracy for the edited images across the three categories changed to 32.16\%, 34.10\%, and 33.74\%, showing an increase in the prediction probability for the medium damage category. Similarly, when edits are applied using a large-sized ROI, the classification accuracy shifted to 31.47\%, 33.69\%, and 34.84\%, indicating that the model increasingly classified the major damage. For Image2, identical experimental results were obtained. The results clearly demonstrate that the edited images, generated using text prompts and ROIs of varying sizes and locations, successfully simulate different levels of building damage. This suggests that the edited images produced through the proposed pipeline have the potential to be effectively utilized in various downstream tasks, mitigating the challenges of obtaining remote sensing images in extreme scenarios.}

\textcolor{black}{In addition to augmenting datasets for disaster change detection and classification tasks}, our proposed image editing method can be effectively utilized in areas such as urban planning and critical target detection. For instance, our image editing method can significantly aid in oil tank detection by generating RSIs that feature oil tanks within diverse geographic backgrounds. Furthermore, it enables the creation of comparative image pairs before and after oil tank construction, supporting tasks like change detection. In our experiment, we employ ROI-prompted image editing to generate RSIs containing oil tanks in various backgrounds, while also controlling the size and number of oil storage tanks in the edited images. The results are presented in Fig. \ref{oiltank}. The experimental results indicate that our method can not only generate oil tanks of varying sizes and quantities in designated areas but also ensure effective integration of the oil tanks with their respective image backgrounds, maintaining consistency in content and details. In addition to generating RSIs featuring oil tanks, our proposed method is capable of producing photovoltaic panels and wind turbines at specified locations, thereby facilitating tasks in specific remote sensing domains. We will present several results generated using the ROI prompt, including site-specific photovoltaic panels and wind turbines. The results are illustrated in Fig. \ref{application}.

\textcolor{black}{In Fig. \ref{application}, the generated RSIs of photovoltaic panels, wind turbine possess high image fidelity, which can alleviate the acquirement for a lot of demand of remote sensing data in specific fields. However, we have to admit that our method is still lacking in the reproduction of high frequency information such as image details and edges, so the generated RSIs containing photovoltaic panels panels are fuzzy inevitably, which should be paid more attention to in the subsequent work.}

\color{black}
\subsection{Ablation Study} 

\begin{figure*}[!ht] \color{black}
\centering
\includegraphics[width = 1.0\textwidth]{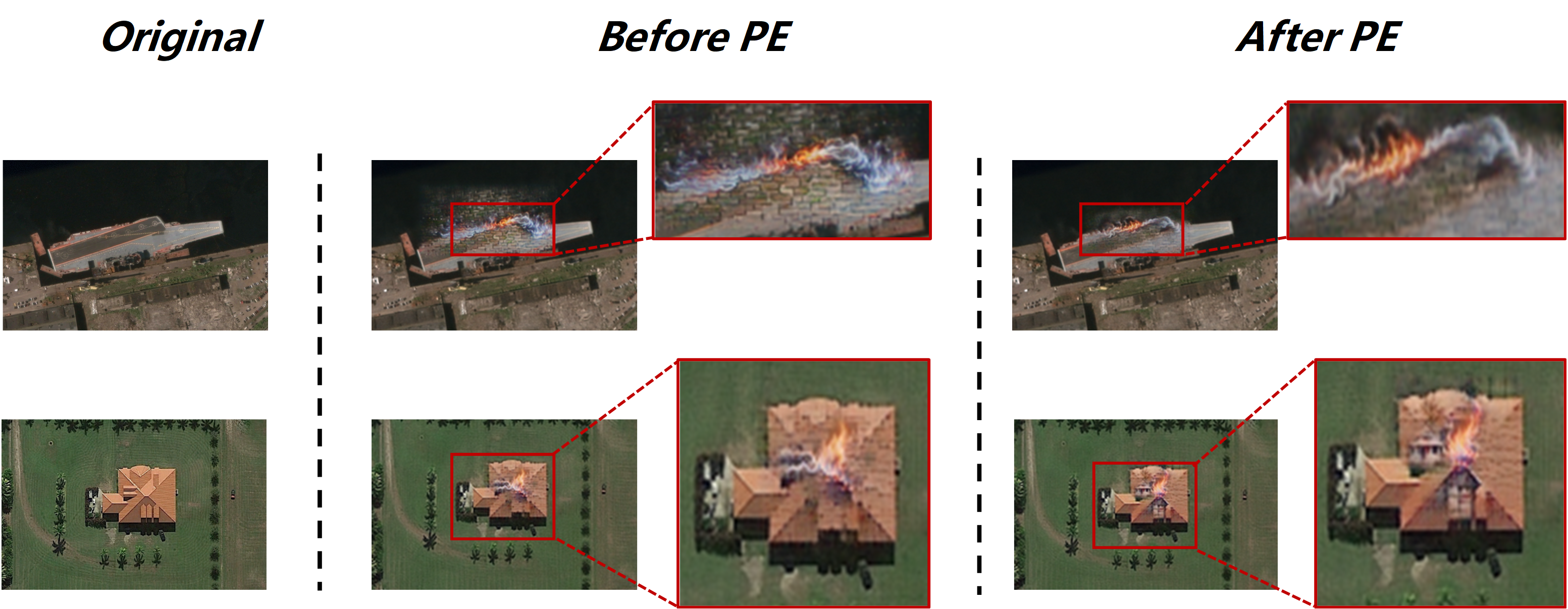}
\caption{The edited images before and after conducting PE when use ``A picture of house/ship in fire'' as the original text prompt.}
\label{ablation}
\end{figure*}

The proposed pipeline emphasizes the importance of two components for image editing, namely multi-scale training strategy based on single image and prompt-guided fine-tuning process combined with PE. In this section, we conduct the ablation experiments. As the multi-scale training strategy is necessary for training DDPM with a single image, we mainly verified the influence of PE on image editing effect. The text guide we use is ``A picture of house/ship in fire'' and the edited images are remote sensing images of ship and house. In the experiment, we use ChatGPT to enhance the same semantics of the text prompt of ``A picture of house/ship in fire''. Corresponding examples of prompt enhancement include:
\begin{enumerate}
\item {A house/ship engulfed in flames}
\item {A burning house/ship}
\item {Flames consuming the house/ship}
\item {A house/ship ablaze}
\item {A house/ship on fire}
\end{enumerate}

Subsequently, following the proposed framework in Fig. \ref{PE}, we employed the text prompts both before and after the PE as guidance to edit the remote sensing images. The experimental results are presented in Fig. \ref{ablation}. Clearly, the post-PE editing yields superior results in terms of preserving image authenticity when compared to using the original text prompt. Specifically, when generating an image of a ship on fire, the absence of PE led to the generation of unrealistic texture details on the deck, with the smoke and flames differing substantially from the actual scene. However, after incorporating PE, the authenticity of the generated images improved markedly. The quantitative indicators also reflect the performance advantages of using PE, as shown in Table \ref{PE_metrics}.

\begin{table}[!ht] \color{black}
\centering
\caption{Quantitative evaluation results of before and after PE. }
\begin{tabular*}{\hsize}{@{}@{\extracolsep{\fill}}ccccccccccccc@{}}
    \hline
    \textbf{} & \textbf{} & Before PE & After PE \\ \hline
    \multirow{2}{*}{CLIP Score (\%)} \textbf{}& House   & 20.04 & 21.38 ($\uparrow$ 1.34) \\ 
    \textbf{} & Ship   & 12.17 & 14.68 ($\uparrow$ 2.51) \\          \hline
    \multirow{2}{*}{Subjective Score} \textbf{}& House  & 3.20  & 4.65 ($\uparrow$ 1.45) \\ 
    \textbf{} & Ship   & 3.72  & 4.44 ($\uparrow$ 0.72)\\ \hline
\end{tabular*}
\label{PE_metrics}
\end{table}

Analyzing the underlying reasons, the semantics of the CLIP model itself are inherently complex. For example, the term ``Fire'' encompasses not only semantics relevant to remote sensing images but also is likely to overlap with artistic or photographic imagery, which can introduce unrealistic texture details in the remote sensing images. Furthermore, the mismatch between textual descriptions and images often results in ``Fire'' being strongly associated with other feature categories, further degrading the quality of the generated edits. With the application of PE, these issues are significantly mitigated. Despite the inherent complexity of the semantics associated with ``Fire'', ChatGPT's advanced text processing and comprehension capabilities allow for the generation of a series of semantically consistent prompts, thereby minimizing the potential for misguidance stemming from reliance on a single text prompt. Guided by these refined prompts, the occurrence of unintended styles or feature categories in the edited images is markedly reduced, thereby enhancing the fidelity of the generated images. The house and ship on fire generated after applying PE not only exhibit highly realistic flames and smoke effects, but also preserve the integrity of the original targets in the image, with minimal alterations. 

\subsection{\textcolor{black}{Special Scenarios}}

\begin{figure*}[!ht] \color{black}
\centering
\includegraphics[width = 0.8\textwidth]{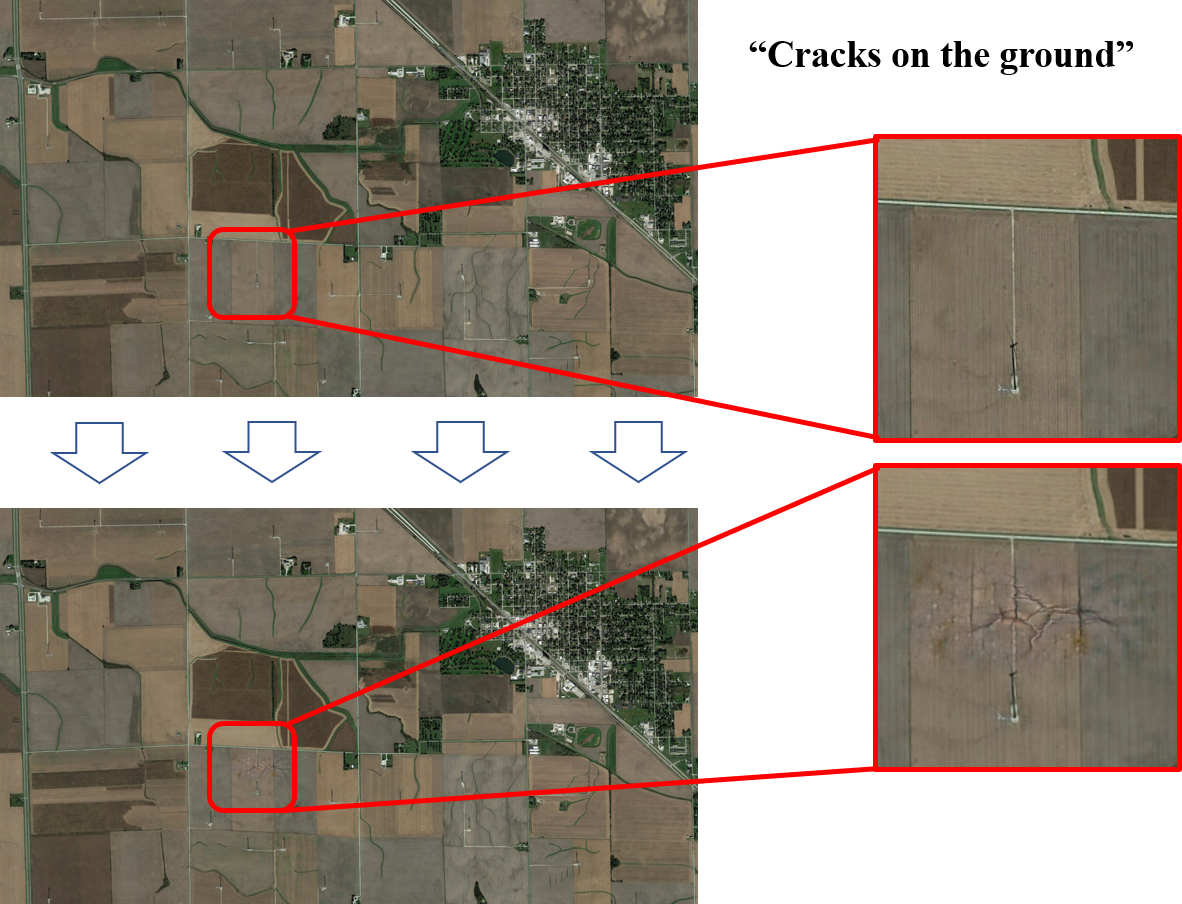}
\caption{The editing result of a specific area within a large-scale RSI with the text prompt ``Cracks on the Ground''. The magnified part of the images show the area before and after image editing. }
\label{large_scene}
\end{figure*}

\textcolor{black}{In addition, we also evaluated the performance of the used image editing methods in large or low-quality scenes. First, as for the large scenes, due to the limitations of the input and computation burden of the model itself, the proposed image editing model can only be used to process RSI slices, not to directly process large-scale remote sensing images. However, since image editing often only occurs in local areas of large-scale remote sensing images, it is possible to edit large-scale RSIs by cropping the large-scale images and then processing the areas of focus specifically. Fig. \ref{large_scene} is an example to illustrate the feasibility of the above process.}

\textcolor{black}{In the above example image, we cropped, edited and stitched to add cracks to parts of a large-scale remote sensing image. However, we also regret to admit that the proposed editing method for large-scale remote sensing images is an expedient measure due to the limitations of the model. In the future, we will further optimize the algorithm in order to achieve both high-resolution and high-efficiency image editing in large-scale remote sensing images.}

\begin{figure*}[!ht] \color{black}
\centering
\includegraphics[width = 1.0\textwidth]{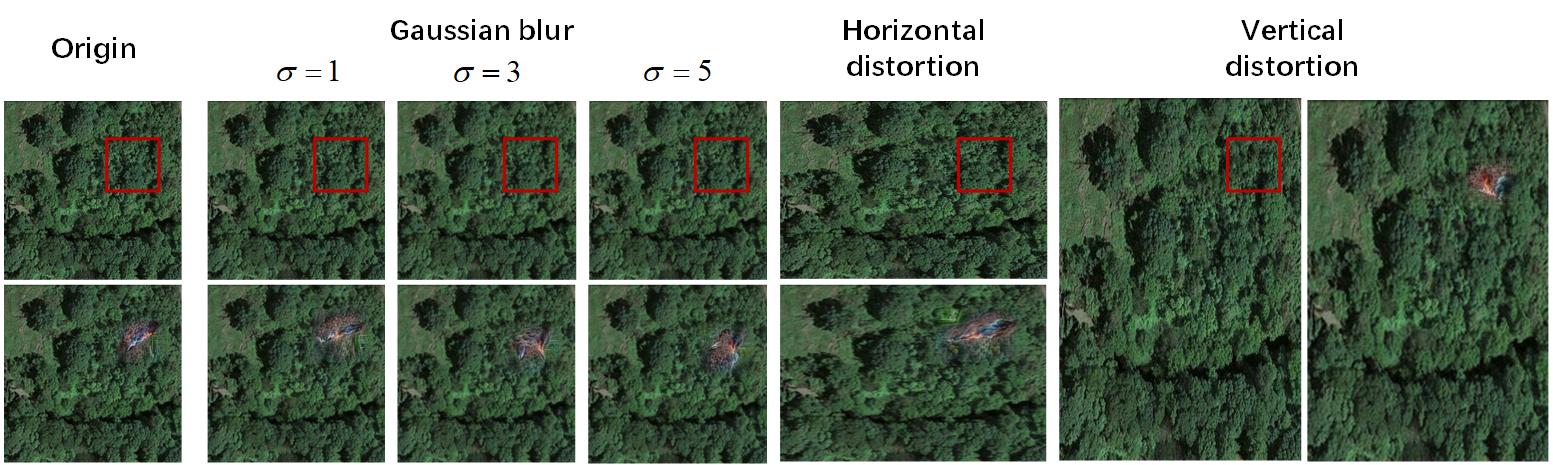}
\caption{Image editing results under different degrees and types of image blurring and distortion. The text prompt is ``A fire in the forest".}
\label{image_blur}
\end{figure*}

\textcolor{black}{At the same time, due to the low-quality issues such as distortion and blurring in RSIs, we conducted experiments on image editing of low-quality RSIs to quantitatively evaluate the impact of the quality of the RSIs themselves on the image editing effect. As to the image blurring, we carried out artificial Gaussian blurring processing on a RSI of a forest area. The Gaussian kernel adopted is fixed at 3, and the mean square deviations are taken as 1, 3 and 5 respectively to construct a set of blurred images of different qualities. As to the image distortion, we adopted two common types, horizontal and vertical, which is manifested as the stretching of the image in a specific direction. We used the text prompt, ``A fire in the forest" as guidance for image editing, and simultaneously employed the CLIP score as the evaluation metric. The experiment results are shown in the Fig. \ref{image_blur} and Table \ref{special_results}. The experimental results show that under the conditions of blurring and distortion, the edited images still maintain a satisfactory editing effect, as demonstrated by the similar scores of the CLIP score. Since the methods based on sinDDM we proposed only receives information from the input image during the pre-training phase, during the editing process, it will inevitably inherit the blurring or distortion of the original image. However, even so, it could be seen from the image that when the basic semantics of the image are not damaged, the results can still maintain good visual effect.} 

\begin{table}[!ht] \color{black}
    \centering
    \caption{The Clip Score results of editing images in scenarios of image blurring and distortion}
    \begin{tabular*}{\hsize}{@{}@{\extracolsep{\fill}}ccccccccccccc@{}}
    \hline
        ~ & \multirow{2}{*}{Original} &\multicolumn{3}{c}{ Blurring} & \multicolumn{2}{c}{Distortion} \\
        ~ & ~ & $\sigma$=1 & $\sigma$=3 & $\sigma$=5 & horizontal  & vertical \\ \hline
        \makecell{Clip Score\\ (100\%)} & 21.72 & 21.89 & 21.03 & 22.13 & 21.67 & 21.40 \\ \hline
    \end{tabular*}
    \label{special_results}
\end{table}

\color{black}
\section{Conclusions}

In this paper, we first address the issue of RSIs editing, a topic that has been relatively overlooked in remote sensing research. We then propose a novel method to achieve stable and controllable prompt-guided RSI editing, utilizing only a single image for training. By employing a multi-scale training strategy and conducting DDPM training with just one sample, we preserve consistency in content and details between the original and edited images and mitigate the challenges associated with scarce training data. Additionally, we leverage VLMs pre-trained on RSI datasets to integrate remote sensing information into the image editing process, using PE to alleviate semantic mismatches between text and image embeddings. Alongside satisfactory visual results, our quantitative evaluation metrics, including CLIP scores and subjective assessments, demonstrate that our method outperforms existing image editing models for RSIs. Furthermore, we apply the generated edited images to support disaster assessment efforts, specifically to evaluate whether a building was destroyed by a tsunami, yielding satisfactory experimental results. 

\textcolor{black}{In future research, our focus will be on three key directions. First, we aim to mitigate high-frequency information degradation during image editing while ensuring task completion, with particular emphasis on bolstering edge preservation and fine-grained detail retention for small objects. Second, we plan to develop adaptive frameworks that can generalize across diverse application scenarios, thereby broadening the applicability of image editing techniques. Concurrently, we will investigate the potential impact of remote sensing image editing on critical domains such as small target detection and multimodal image fusion, exploring how advanced editing capabilities can inform and enhance these related fields.}

\bibliographystyle{ieeetr}
\bibliography{ref.bib}

\begin{thebibliography}{10}

\bibitem{10380634}
X.~Xing, B.~Yu, C.~Kang, B.~Huang, J.~Gong, and Y.~Liu, ``The synergy between remote sensing and social sensing in urban studies: Review and perspectives,'' {\em IEEE Geosci. Remote Sens. Mag.}, vol.~12, no.~1, pp.~108--137, 2024.

\bibitem{10542538}
K.~Chen, B.~Chen, C.~Liu, W.~Li, Z.~Zou, and Z.~Shi, ``Rsmamba: Remote sensing image classification with state space model,'' {\em IEEE Geosci. Remote Sens. Lett.}, vol.~21, pp.~1--5, 2024.

\bibitem{NEURIPS2018_73e5080f}
J.~Klys, J.~Snell, and R.~Zemel, ``Learning latent subspaces in variational autoencoders,'' in {\em Proc. Adv. Neural Inf. Process. Syst.}, vol.~31, pp.~6444--6454, 2018.

\bibitem{10287367}
D.~Li, X.~Nie, R.~Gong, X.~Lin, and H.~Yu, ``Multi-branch gan-based abnormal events detection via context learning in surveillance videos,'' {\em IEEE Trans. Circuits Syst. Video Technol.}, vol.~34, no.~5, pp.~3439--3450, 2024.

\bibitem{9316788}
J.~Song, J.~Li, H.~Chen, and J.~Wu, ``Mapgen-gan: A fast translator for remote sensing image to map via unsupervised adversarial learning,'' {\em IEEE J. Sel. Top. Appl. Earth Obs. Remote Sens.}, vol.~14, pp.~2341--2357, 2021.

\bibitem{9250622}
X.~Pan, J.~Zhao, and J.~Xu, ``Conditional generative adversarial network-based training sample set improvement model for the semantic segmentation of high-resolution remote sensing images,'' {\em IEEE Trans. Geosci. Remote Sens.}, vol.~59, no.~9, pp.~7854--7870, 2021.

\bibitem{9858033}
J.~Qin, Z.~Liu, L.~Ran, R.~Xie, J.~Tang, and Z.~Guo, ``A target sar image expansion method based on conditional wasserstein deep convolutional gan for automatic target recognition,'' {\em IEEE J. Sel. Top. Appl. Earth Obs. Remote Sens.}, vol.~15, pp.~7153--7170, 2022.

\bibitem{10025584}
Y.~Sun, Y.~Wang, L.~Hu, Y.~Huang, H.~Liu, S.~Wang, and C.~Zhang, ``Attribute-guided generative adversarial network with improved episode training strategy for few-shot sar image generation,'' {\em IEEE J. Sel. Top. Appl. Earth Obs. Remote Sens.}, vol.~16, pp.~1785--1801, 2023.

\bibitem{9531488}
N.~Lv, H.~Ma, C.~Chen, Q.~Pei, Y.~Zhou, F.~Xiao, and J.~Li, ``Remote sensing data augmentation through adversarial training,'' {\em IEEE J. Sel. Top. Appl. Earth Obs. Remote Sens.}, vol.~14, pp.~9318--9333, 2021.

\bibitem{8677274}
K.~Jiang, Z.~Wang, P.~Yi, G.~Wang, T.~Lu, and J.~Jiang, ``Edge-enhanced gan for remote sensing image superresolution,'' {\em IEEE Trans. Geosci. Remote Sens.}, vol.~57, no.~8, pp.~5799--5812, 2019.

\bibitem{10403859}
H.~Li, W.~Deng, Q.~Zhu, Q.~Guan, and J.~Luo, ``Local-global context-aware generative dual-region adversarial networks for remote sensing scene image super-resolution,'' {\em IEEE Trans. Geosci. Remote Sens.}, vol.~62, pp.~1--14, 2024.

\bibitem{10763472}
J.~Sui, X.~Ma, X.~Zhang, M.-O. Pun, and H.~Wu, ``Adaptive semantic-enhanced denoising diffusion probabilistic model for remote sensing image super-resolution,'' {\em IEEE J. Sel. Top. Appl. Earth Obs. Remote Sens.}, vol.~18, pp.~892--906, 2025.

\bibitem{9323085}
K.~Doi, K.~Sakurada, M.~Onishi, and A.~Iwasaki, ``Gan-based sar-to-optical image translation with region information,'' in {\em Proc. IEEE Int. Geosci. Remote Sens. Symp.}, pp.~2069--2072, 2020.

\bibitem{10707182}
Y.~Zhang, R.~Fan, P.~Duan, J.~Dong, and Z.~Lei, ``Dcdgan-stf: A multiscale deformable convolution distillation gan for remote sensing image spatiotemporal fusion,'' {\em IEEE J. Sel. Top. Appl. Earth Obs. Remote Sens.}, vol.~17, pp.~19436--19450, 2024.

\bibitem{10103165}
B.~Wang, P.~Chen, and G.~Zhang, ``Simulation of gpr b-scan data based on dense generative adversarial network,'' {\em IEEE J. Sel. Top. Appl. Earth Obs. Remote Sens.}, vol.~16, pp.~3938--3944, 2023.

\bibitem{10382538}
R.~Zhang, Z.~Cao, S.~Yang, L.~Si, H.~Sun, L.~Xu, and F.~Sun, ``Cognition-driven structural prior for instance-dependent label transition matrix estimation,'' {\em IEEE Trans. Neural Netw. Learn. Syst.}, pp.~1--14, 2024.

\bibitem{10716600}
Y.~Kang, J.~Wu, Q.~Liu, J.~Yue, and L.~Fang, ``Trans-diff: Heterogeneous domain adaptation for remote sensing segmentation with transfer diffusion,'' {\em IEEE J. Sel. Top. Appl. Earth Obs. Remote Sens.}, vol.~17, pp.~18413--18426, 2024.

\bibitem{9481173}
Y.~Zhao, S.~Shen, J.~Hu, Y.~Li, and J.~Pan, ``Cloud removal using multimodal gan with adversarial consistency loss,'' {\em IEEE Geosci. Remote Sens. Lett.}, vol.~19, pp.~1--5, 2022.

\bibitem{10445007}
J.~Zhang, J.~Huang, S.~Jin, and S.~Lu, ``Vision-language models for vision tasks: A survey,'' {\em IEEE Trans. Pattern Anal. Mach. Intell.}, vol.~46, no.~8, pp.~5625--5644, 2024.

\bibitem{radford2021learning}
A.~Radford, J.~W. Kim, C.~Hallacy, A.~Ramesh, G.~Goh, S.~Agarwal, G.~Sastry, A.~Askell, P.~Mishkin, J.~Clark, G.~Krueger, and I.~Sutskever, ``Learning transferable visual models from natural language supervision,'' {\em arXiv preprint arXiv:2103.00020}, 2021.

\bibitem{ho2020denoising}
J.~Ho, A.~Jain, and P.~Abbeel, ``Denoising diffusion probabilistic models,'' {\em arXiv preprint arXiv:2006.11239}, 2020.

\bibitem{xia2021tedigan}
W.~Xia, Y.~Yang, J.-H. Xue, and B.~Wu, ``Tedigan: Text-guided diverse face image generation and manipulation,'' in {\em Proc. IEEE Conf. Comput. Vis. Pattern Recognit.}, pp.~2256--2265, 2021.

\bibitem{Patashnik_2021_ICCV}
O.~Patashnik, Z.~Wu, E.~Shechtman, D.~Cohen-Or, and D.~Lischinski, ``Styleclip: Text-driven manipulation of stylegan imagery,'' in {\em Proc. IEEE Int. Conf. Comput. Vis.}, pp.~2085--2094, 2021.

\bibitem{huang2024}
Y.~Huang, J.~Huang, Y.~Liu, M.~Yan, J.~Lv, J.~Liu, W.~Xiong, H.~Zhang, S.~Chen, and L.~Cao, ``Diffusion model-based image editing: A survey,'' {\em arXiv preprint: arXiv 2402.17525}, 2024.

\bibitem{10387416}
N.~Huang, Y.~Zhang, F.~Tang, C.~Ma, H.~Huang, W.~Dong, and C.~Xu, ``Diffstyler: Controllable dual diffusion for text-driven image stylization,'' {\em IEEE Trans. Neural Networks Learn. Syst.}, pp.~1--14, 2024.

\bibitem{Wang_2023_ICCV}
Z.~Wang, L.~Zhao, and W.~Xing, ``Stylediffusion: Controllable disentangled style transfer via diffusion models,'' in {\em Proc. IEEE Int. Conf. Comput. Vis.}, pp.~7677--7689, October 2023.

\bibitem{Mokady_2023_CVPR}
R.~Mokady, A.~Hertz, K.~Aberman, Y.~Pritch, and D.~Cohen-Or, ``Null-text inversion for editing real images using guided diffusion models,'' in {\em Proc. IEEE Conf. Comput. Vis. Pattern Recog.}, pp.~6038--6047, June 2023.

\bibitem{Zhang_CVPR}
Y.~Zhang, N.~Huang, F.~Tang, H.~Huang, C.~Ma, W.~Dong, and C.~Xu, ``Inversion-based style transfer with diffusion models,'' in {\em Proc. IEEE Conf. Comput. Vis. Pattern Recog.}, pp.~10146--10156, June 2023.

\bibitem{Wallace_2023_CVPR}
B.~Wallace, A.~Gokul, and N.~Naik, ``Edict: Exact diffusion inversion via coupled transformations,'' in {\em Proc. IEEE Conf. Comput. Vis. Pattern Recog.}, pp.~22532--22541, June 2023.

\bibitem{NEURIPS2023_3469b211}
D.~Epstein, A.~Jabri, B.~Poole, A.~Efros, and A.~Holynski, ``Diffusion self-guidance for controllable image generation,'' in {\em Proc. Adv. Neural Inf. Process. Syst.}, vol.~36, pp.~16222--16239, 2023.

\bibitem{Kawar_2023_CVPR}
B.~Kawar, S.~Zada, O.~Lang, O.~Tov, H.~Chang, T.~Dekel, I.~Mosseri, and M.~Irani, ``Imagic: Text-based real image editing with diffusion models,'' in {\em Proc. IEEE Conf. Comput. Vis. Pattern Recognit.}, pp.~6007--6017, June 2023.

\bibitem{8672156}
P.~Ghamisi, B.~Rasti, N.~Yokoya, Q.~Wang, B.~Hofle, L.~Bruzzone, F.~Bovolo, M.~Chi, K.~Anders, R.~Gloaguen, P.~M. Atkinson, and J.~A. Benediktsson, ``Multisource and multitemporal data fusion in remote sensing: A comprehensive review of the state of the art,'' {\em IEEE Geosci. Remote Sens. Mag.}, vol.~7, no.~1, pp.~6--39, 2019.

\bibitem{Li_2024_CVPR}
Z.~Li, F.~Lu, J.~Zou, L.~Hu, and H.~Zhang, ``Generalized few-shot meets remote sensing: Discovering novel classes in land cover mapping via hybrid semantic segmentation framework,'' in {\em Proc. IEEE Conf. Comput. Vis. Pattern Recog.}, pp.~2744--2754, June 2024.

\bibitem{10356128}
L.~Si, H.~Dong, W.~Qiang, Z.~Song, B.~Du, J.~Yu, and F.~Sun, ``A trusted generative-discriminative joint feature learning framework for remote sensing image classification,'' {\em IEEE Trans. Geosci. Remote Sens.}, vol.~62, p.~5601814, 2024.

\bibitem{9393473}
L.~Zhang and Y.~Liu, ``Remote sensing image generation based on attention mechanism and vae-msgan for roi extraction,'' {\em IEEE Geosci. Remote Sens. Lett.}, vol.~19, pp.~1--5, 2022.

\bibitem{9386248}
H.~Chen, W.~Li, and Z.~Shi, ``Adversarial instance augmentation for building change detection in remote sensing images,'' {\em IEEE Trans. Geosci. Remote Sens.}, vol.~60, pp.~1--16, 2022.

\bibitem{9328132}
R.~Dong, L.~Zhang, and H.~Fu, ``Rrsgan: Reference-based super-resolution for remote sensing image,'' {\em IEEE Trans. Geosci. Remote Sens.}, vol.~60, pp.~1--17, 2022.

\bibitem{10353979}
Y.~Xiao, Q.~Yuan, K.~Jiang, J.~He, X.~Jin, and L.~Zhang, ``Ediffsr: An efficient diffusion probabilistic model for remote sensing image super-resolution,'' {\em IEEE Trans. Geosci. Remote Sens.}, vol.~62, pp.~1--14, 2024.

\bibitem{Peebles_2023_ICCV}
W.~Peebles and S.~Xie, ``Scalable diffusion models with transformers,'' in {\em Proc. IEEE Int. Conf. Comput. Vis.}, pp.~4195--4205, October 2023.

\bibitem{NEURIPS2021_b578f2a5}
D.~Kingma, T.~Salimans, B.~Poole, and J.~Ho, ``Variational diffusion models,'' in {\em Proc. Adv. Neural Inf. Process. Syst.}, vol.~34, pp.~21696--21707, 2021.

\bibitem{10633292}
S.~Kang, S.~Gao, W.~Wu, X.~Wang, S.~Wang, and G.~Qiu, ``Image intrinsic components guided conditional diffusion model for low-light image enhancement,'' {\em IEEE Trans. Circuits Syst. Video Technol.}, vol.~34, no.~12, pp.~13244--13256, 2024.

\bibitem{10480695}
H.~Liu, L.~He, M.~Zhang, and F.~Li, ``Vadiffusion: Compressed domain information guided conditional diffusion for video anomaly detection,'' {\em IEEE Trans. Circuits Syst. Video Technol.}, vol.~34, no.~9, pp.~8398--8411, 2024.

\bibitem{pmlr-v202-kulikov23a}
V.~Kulikov, S.~Yadin, M.~Kleiner, and T.~Michaeli, ``{S}in{DDM}: A single image denoising diffusion model,'' in {\em Proc. Int. Conf. Mach. Learn.}, vol.~202, pp.~17920--17930, 2023.

\bibitem{10542240}
Z.~Chen, X.~Xu, Y.~Yan, Y.~Pan, W.~Zhu, W.~Wu, B.~Dai, and X.~Yang, ``Hyperstyle3d: Text-guided 3d portrait stylization via hypernetworks,'' {\em IEEE Trans. Circuits Syst. Video Technol.}, vol.~34, no.~10, pp.~9997--10010, 2024.

\bibitem{10154005}
Z.~Yang, T.~Chu, X.~Lin, E.~Gao, D.~Liu, J.~Yang, and C.~Wang, ``Eliminating contextual prior bias for semantic image editing via dual-cycle diffusion,'' {\em IEEE Trans. Circuits Syst. Video Technol.}, vol.~34, no.~2, pp.~1316--1320, 2024.

\bibitem{9716741}
K.~Han, Y.~Wang, H.~Chen, X.~Chen, J.~Guo, Z.~Liu, Y.~Tang, A.~Xiao, C.~Xu, Y.~Xu, Z.~Yang, Y.~Zhang, and D.~Tao, ``A survey on vision transformer,'' {\em IEEE Trans. Pattern Anal. Mach. Intell.}, vol.~45, no.~1, pp.~87--110, 2023.

\bibitem{Khattak_2023_CVPR}
M.~U. Khattak, H.~Rasheed, M.~Maaz, S.~Khan, and F.~S. Khan, ``Maple: Multi-modal prompt learning,'' in {\em Proc. IEEE Conf. Comput. Vis. Pattern Recog.}, pp.~19113--19122, June 2023.

\bibitem{Liu_2023_CVPR}
Y.~Liu, Y.~Lu, H.~Liu, Y.~An, Z.~Xu, Z.~Yao, B.~Zhang, Z.~Xiong, and C.~Gui, ``Hierarchical prompt learning for multi-task learning,'' in {\em Proc. IEEE Conf. Comput. Vis. Pattern Recog.}, pp.~10888--10898, June 2023.

\bibitem{Park_2024_CVPR}
J.~Park, J.~Ko, and H.~J. Kim, ``Prompt learning via meta-regularization,'' in {\em Proc. IEEE Conf. Comput. Vis. Pattern Recog.}, pp.~26940--26950, June 2024.

\bibitem{7907303}
G.-S. Xia, J.~Hu, F.~Hu, B.~Shi, X.~Bai, Y.~Zhong, L.~Zhang, and X.~Lu, ``Aid: A benchmark data set for performance evaluation of aerial scene classification,'' {\em IEEE Trans. Geosci. Remote Sens.}, vol.~55, no.~7, pp.~3965--3981, 2017.

\bibitem{Chen2022MSSDet}
W.~Chen, B.~Han, Z.~Yang, and X.~Gao, ``Mssdet: Multi-scale ship-detection framework in optical remote-sensing images and new benchmark,'' {\em Remote Sens.}, vol.~14, no.~21, p.~5460, 2022.

\bibitem{remoteclip}
F.~Liu, D.~Chen, Z.~Guan, X.~Zhou, J.~Zhu, Q.~Ye, L.~Fu, and J.~Zhou, ``Remoteclip: {A} vision language foundation model for remote sensing,'' {\em arXiv preprint: arXiv 2306.11029}, 2023.

\bibitem{DBLP:journals/corr/abs-2104-08718}
J.~Hessel, A.~Holtzman, M.~Forbes, R.~L. Bras, and Y.~Choi, ``Clipscore: {A} reference-free evaluation metric for image captioning,'' {\em arXiv preprint: arXiv 2104.08718}, 2021.

\bibitem{rombach2021highresolution}
R.~Rombach, A.~Blattmann, D.~Lorenz, P.~Esser, and B.~Ommer, ``High-resolution image synthesis with latent diffusion models,'' 2021.

\bibitem{kandinsky2.1}
A.~Shakhmatov, A.~Razzhigaev, A.~Nikolich, V.~Arkhipkin, I.~Pavlov, A.~Kuznetsov, and D.~Dimitrov, ``Kandinsky 2.1,'' 2023.

\bibitem{fu2024}
T.-J. Fu, W.~Hu, X.~Du, W.~Y. Wang, Y.~Yang, and Z.~Gan, ``Guiding instruction-based image editing via multimodal large language models,'' 2024.

\bibitem{zhang2025incontexteditenablinginstructional}
Z.~Zhang, J.~Xie, Y.~Lu, Z.~Yang, and Y.~Yang, ``In-context edit: Enabling instructional image editing with in-context generation in large scale diffusion transformer,'' 2025.

\bibitem{7986759}
A.~Fujita, K.~Sakurada, T.~Imaizumi, R.~Ito, S.~Hikosaka, and R.~Nakamura, ``Damage detection from aerial images via convolutional neural networks,'' in {\em Proc. IAPR Int. Conf. Mach. Vis. Appl.}, pp.~5--8, 2017.

\bibitem{Rahnemoonfar2023}
M.~Rahnemoonfar, T.~Chowdhury, and R.~Murphy, ``Rescuenet: A high resolution uav semantic segmentation dataset for natural disaster damage assessment,'' {\em Scientific Data}, vol.~10, no.~1, p.~913, 2023.

\end{thebibliography}

\begin{IEEEbiography}[{\includegraphics[width=1in,height=1.25in,clip,keepaspectratio]{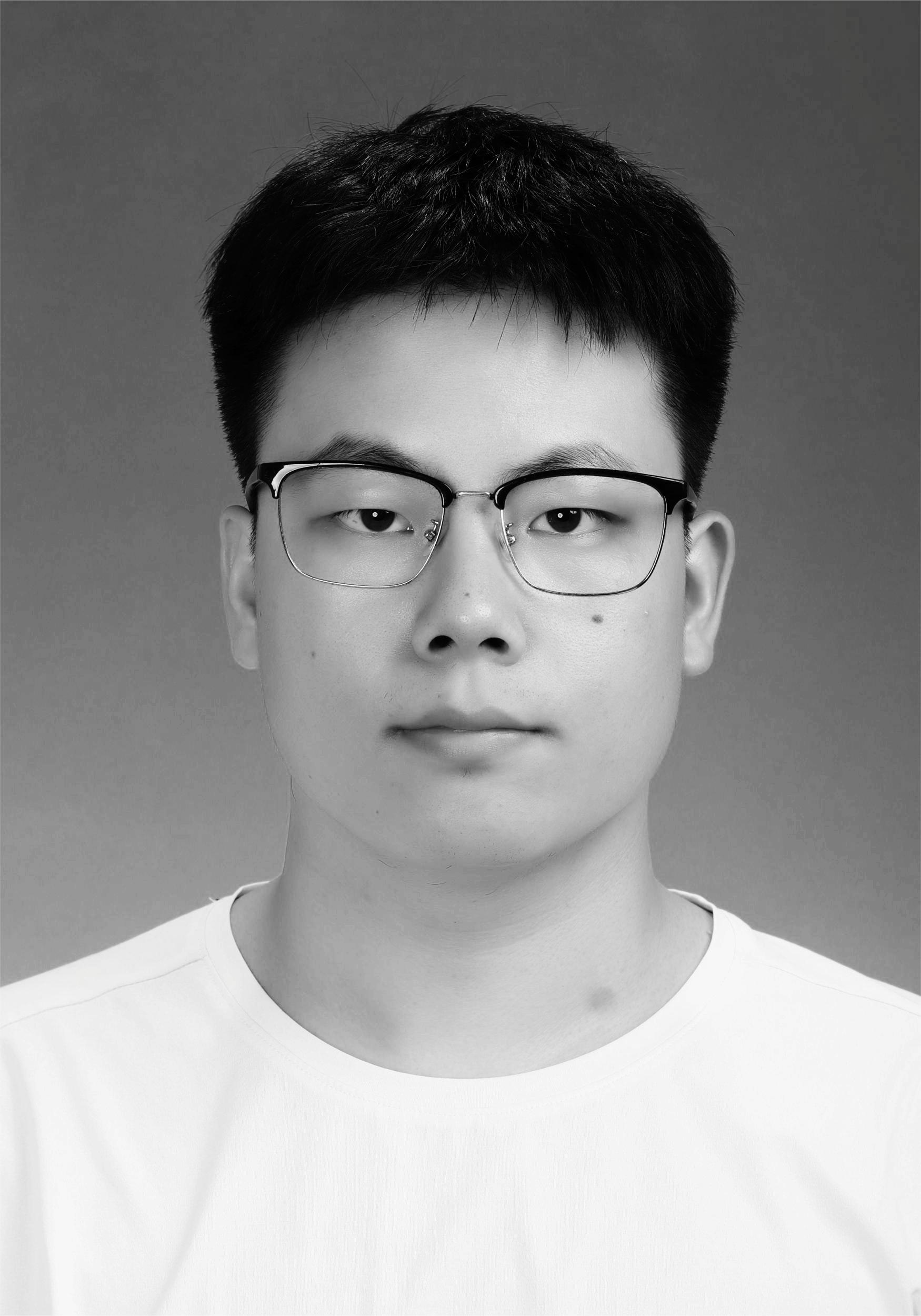}}]{Fangzhou Han} received the B.E. and M. S. degree in information and communication engineering from Harbin Institute of Technology, Harbin, China. He is currently pursuing the Ph.D. degree with Shenzhen Graduate School, Harbin Institute of Technology, Shenzhen, China. His research interests include deep learning, remote sensing and multimodal learning.
\end{IEEEbiography}
\vspace{-1.0cm}
\begin{IEEEbiography}[{\includegraphics[width=1in,height=1.25in,clip,keepaspectratio]{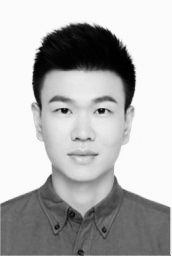}}]{Lingyu Si} received the M.S. degree from University of Bristol, in 2018, and the Ph.D. degree from University of Chinese Academy of Sciences, Beijing, China, in 2024. His research interests include transfer learning and self-supervised learning.
\end{IEEEbiography}
\vspace{-1.0cm}
\begin{IEEEbiography}[{\includegraphics[width=1in,height=1.25in,clip,keepaspectratio]{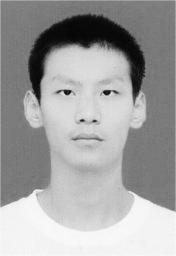}}]
{Hongwei Dong} (Member, IEEE) received the M.S. degree from China Agricultural University, Beijing, China, in 2018, and the Ph.D. degree from Harbin Institute of Technology, Harbin, China, in 2022. He is currently a research fellow with the National Key Laboratory of Space Integrated Information System, Institute of Software, Chinese Academy of Sciences. His current research interests include applied mathematics, optimization methods, and machine learning.
\end{IEEEbiography}
\vspace{-1.0cm}

\begin{IEEEbiography}[{\includegraphics[width=1in,height=1.25in,clip,keepaspectratio]{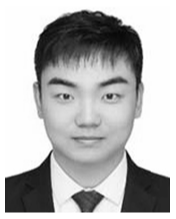}}]
{Zhizhuo Jiang} received the B.S. degree in Electronics and Information Engineering and the M.S. degree in Information and Communication Engineering from the Harbin Institute of Technology, Harbin, China, in 2014 and 2016, respectively, and the Ph.D. degree in Information and Communication Engineering from Tsinghua University, Beijing, China. He is currently a Post-Doctoral Fellow at the Tsinghua Shenzhen International Graduate School, Shenzhen, China. His main research interests include target detection, multimodal data fusion, deep learning, and signal processing.
\end{IEEEbiography}
\vspace{-1.0cm}

\begin{IEEEbiography}[{\includegraphics[width=1in,height=1.25in,clip,keepaspectratio]{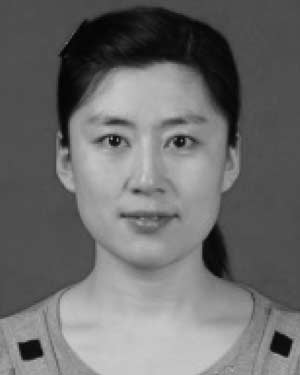}}]{Lamei Zhang} (Senior Member, IEEE) received the B.S., M.Sc., and Ph.D. degrees in information and communication engineering from Harbin Institute of Technology, Harbin, China, in 2004, 2006, and 2010, respectively. She is currently an Associate Professor with the Department of Information Engineering, Harbin Institute of Technology. Her research interests include RSIs processing, information extraction and interpretation of high-resolution synthetic aperture radar, polarimetric SAR, and polarimetric SAR interferometry. Dr. Zhang serves as the Secretary for the IEEE Harbin Education Section.
\end{IEEEbiography}
\vspace{-1.0cm}
\begin{IEEEbiography}[{\includegraphics[width=1in,height=1.25in,clip,keepaspectratio]{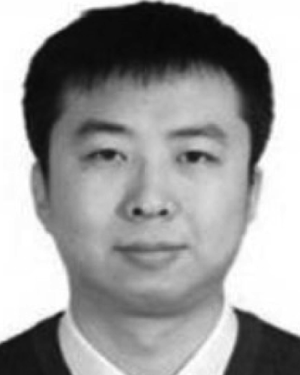}}]{Hao Chen} (Member, IEEE) received the B.S., M.S., and Ph.D. degrees from Harbin Institute of Technology, Harbin, China, in 2001, 2003, and 2008, respectively. Since 2004, he has been with the School of Electronics and Information Engineering, Harbin Institute of Technology. He is currently a Professor. His main research interests include RSI processing and image compression.
\end{IEEEbiography}

\begin{IEEEbiography}[{\includegraphics[width=1in,height=1.25in,clip,keepaspectratio]{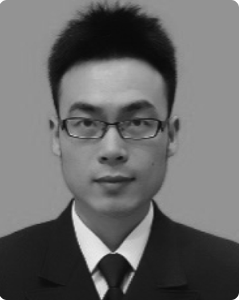}}]
{Yu Liu} (Member, IEEE) received the B.S. and Ph.D. degrees in information and communication engineering from Naval Aviation University, Yantai, China, in 2008 and 2014, respectively. Since 2014, he has been with the faculty of Naval Aviation University, where he is currently a Professor with the Research Institute of Information Fusion. From 2016 to 2018, he was a Post-Doctoral Researcher with the Department of Information and Communication Engineering, Beihang University, Beijing, China. He is now a visiting professor at Tsinghua University. His research interests include multi-sensor fusion and remote data processing.
\end{IEEEbiography}
\end{document}